\documentclass[10pt,twocolumn,letterpaper]{article}

\usepackage{iccv}
\usepackage{times}
\usepackage{epsfig}
\usepackage{graphicx}
\usepackage{amsmath}
\usepackage{amssymb}
\usepackage{dsfont}
\usepackage{bbm}
\usepackage{booktabs}
\usepackage{multirow}

\usepackage[pagebackref=true,breaklinks=true,letterpaper=true,colorlinks,bookmarks=false]{hyperref}

\iccvfinalcopy %

\def\iccvPaperID{1012} %

\newcommand{\mb}[1]{\mathbf{#1}}
\newcommand{\bs}[1]{\boldsymbol{#1}}
\newcommand{\mname}{AgentFormer}

\ificcvfinal\fi

\begin{document}

\title{\mname: Agent-Aware Transformers for\\Socio-Temporal Multi-Agent Forecasting}

\author{
Ye Yuan\textsuperscript{1} \qquad Xinshuo Weng\textsuperscript{1} \qquad Yanglan Ou\textsuperscript{2} \qquad Kris Kitani\textsuperscript{1}\\[1mm]
\textsuperscript{1}Carnegie Mellon University \qquad \textsuperscript{2}Penn State University\\[1mm]
{ \url{https://www.ye-yuan.com/agentformer}} \\
}

\maketitle
\ificcvfinal\thispagestyle{empty}\fi

\begin{abstract}
\vspace{-3mm}
    Predicting accurate future trajectories of multiple agents is essential for autonomous systems but is challenging due to the complex interaction between agents and the uncertainty in each agent's future behavior. Forecasting multi-agent trajectories requires modeling two key dimensions:
    (1)~\textbf{time dimension}, where we model the influence of past agent states over future states; (2)~\textbf{social dimension}, where we model how the state of each agent affects others.
    Most prior methods model these two dimensions separately, e.g., first using a temporal model to summarize features over time for each agent independently and then modeling the interaction of the summarized features with a social model. This approach is suboptimal since independent feature encoding over either the time or social dimension can result in a loss of information. Instead, we would prefer a method that allows an agent's state at one time to \textbf{directly} affect another agent's state at a future time.
    To this end, we propose a new Transformer, termed \mname, that simultaneously models the time and social dimensions. The model leverages a sequence representation of multi-agent trajectories by flattening trajectory features across time and agents. Since standard attention operations disregard the agent identity of each element in the sequence, \mname\ uses a novel agent-aware attention mechanism that preserves agent identities by attending to elements of the same agent differently than elements of other agents. Based on \mname, we propose a stochastic multi-agent trajectory prediction model that can attend to features of any agent at any previous timestep when inferring an agent's future position. The latent intent of all agents is also jointly modeled, allowing the stochasticity in one agent's behavior to affect other agents. Extensive experiments show that our method substantially improves the state of the art on well-established pedestrian and autonomous driving datasets.
\end{abstract}

\begin{figure}
    \centering
    \includegraphics[width=\linewidth]{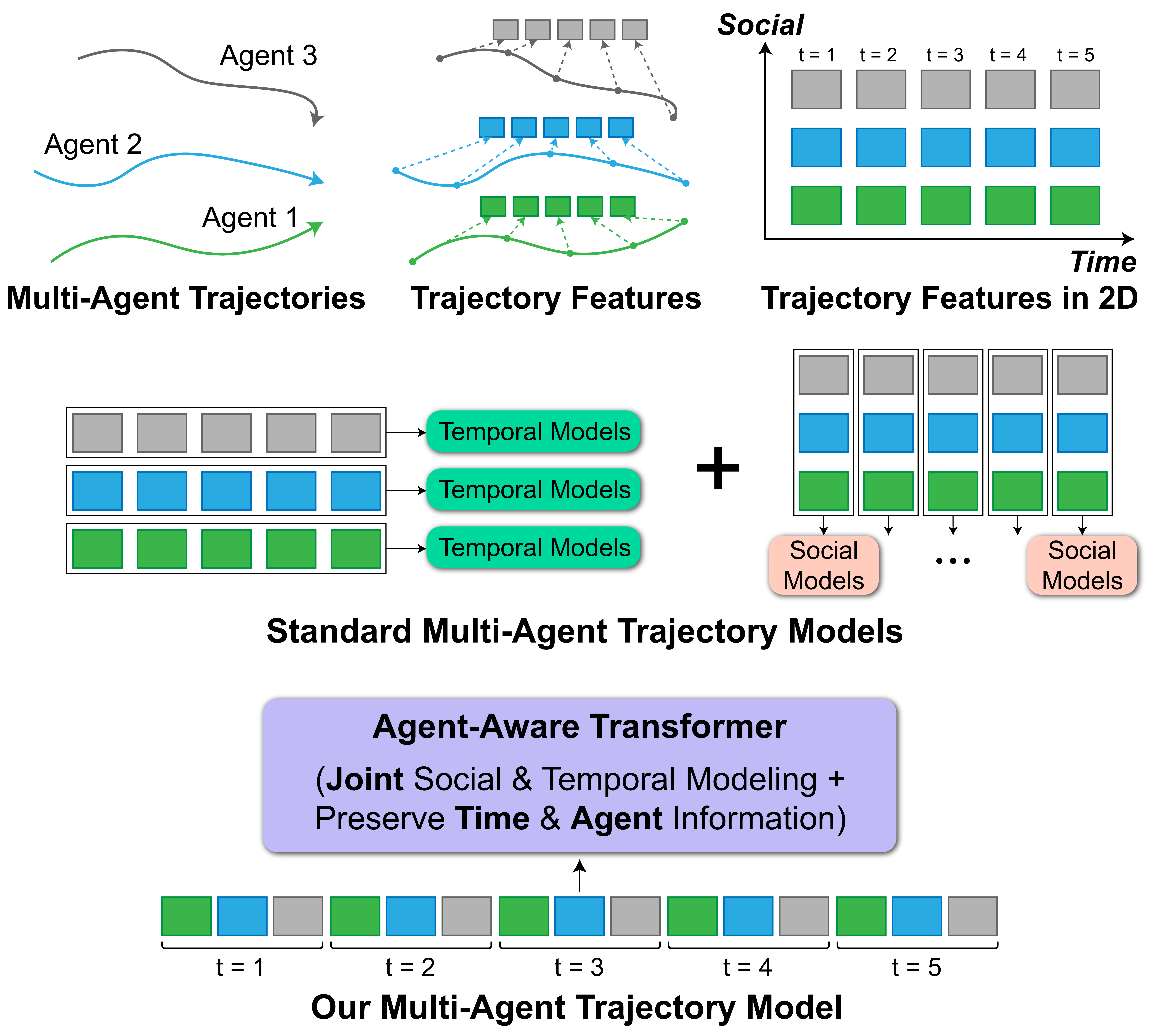}
    \caption{Different from standard approaches that model multi-agent trajectories in the time and social dimensions separately, our \mname\ allows for joint modeling of the time and social dimensions while preserving time and agent information. }
    \label{fig:teaser}
    \vspace{-4mm}
\end{figure}

\vspace{-5mm}
\section{Introduction}
\vspace{-1mm}
\label{sec:intro}
The safe planning of autonomous systems such as self-driving vehicles requires forecasting accurate future trajectories of surrounding agents (\eg, pedestrians, vehicles). However, multi-agent trajectory forecasting is challenging since the social interaction between agents, \ie, behavioral influence of an agent on others, is a complex process. The problem is further complicated by the uncertainty of each agent's future behavior, \ie, each agent has its latent intent unobserved by the system (\eg, turning left or right) that governs its future trajectory and in turn affects other agents. Therefore, a good multi-agent trajectory forecasting method should effectively model (1) the complex social interaction between agents and (2) the latent intent of each agent's future behavior and its social influence on other agents. 

Multi-agent social interaction modeling involves two key dimensions as illustrated in Fig.~\ref{fig:teaser}\,(Top): (1)~\textbf{\emph{time dimension}}, where we model how past agent states (positions and velocities) influence future agent states; (2)~\textbf{\emph{social dimension}}, where we model how each agent's state affects the state of other agents. Most prior multi-agent trajectory forecasting methods model these two dimensions separately (see Fig.~\ref{fig:teaser}\,(Middle)). Approaches like~\cite{kosaraju2019social,alahi2016social,gupta2018social} first use temporal models (\eg, LSTMs~\cite{hochreiter1997long} or Transformers~\cite{vaswani2017attention}) to summarize trajectory features over time for each agent independently and then input the summarized temporal features to social models (\eg, graph neural networks~\cite{kipf2016semi}) to capture social interaction between agents. Alternatively, methods like~\cite{salzmann2020trajectron++,huang2019stgat} first use social models to produce social features for each agent at each independent timestep and then apply temporal models over the social features. In this work, we argue that modeling the time and social dimensions separately can be suboptimal since the independent feature encoding over either the time or social dimension is not informed by features across the other dimension, and the encoded features may not contain the necessary information for modeling the other dimension.

To tackle this problem, we propose a new Transformer model, termed \mname, that simultaneously learns representations from both the time and social dimensions. \mbox{\mname}\ allows an agent's state at one time to affect another agent's state at a future time \emph{directly} instead of through intermediate features encoded over one dimension. As Transformers require sequences as input, we leverage a sequence representation of multi-agent trajectories by flattening trajectory features across time and agents (see Fig.~\ref{fig:teaser}\,(Bottom)). However, directly applying standard Transformers to these multi-agent sequences will result in a loss of \emph{time} and \emph{agent} information since standard attention operations discard the timestep and agent identity associated with each element in the sequence. We solve the loss of time information using a time encoder that appends a timestamp feature to each element. However, the loss of agent identity is a more complicated problem: unlike time, there is no innate ordering between agents, and assigning an agent index-based encoding will break the required permutation invariance of agents and create artificial dependencies on agent indices in the model. Instead, we propose a novel agent-aware attention mechanism to preserve agent information. Specifically, agent-aware attention generates two sets of keys and queries via different linear transformations; one set of keys and queries is used to compute inter-agent attention  (agent to agent) while the other set is designated for intra-agent attention (agent to itself). This design allows agent-aware attention to attend to elements of the same agent differently than elements of other agents, thus keeping the notion of agent identity. Agent-aware attention can be implemented efficiently via masked operations. Furthermore, \mname\ can also encode rule-based connectivity between agents (\eg, based on distance) by masking out the attention weights between unconnected agents.

Based on \mname, which allows us to model social interaction effectively, we propose a multi-agent trajectory prediction framework that also models the social influence of each agent's future trajectory on other agents. The probabilistic formulation of the model follows the conditional variational autoencoder (CVAE~\cite{kingma2013auto}) where we model the generative future trajectory distribution conditioned on context (\eg, past trajectories, semantic maps). We introduce a latent code for each agent to represent its latent intent. To model the social influence of each agent's future behavior (governed by latent intent) on other agents, the latent codes of all agents are jointly inferred from the future trajectories of all agents during training, and they are also jointly used by a trajectory decoder to output socially-aware multi-agent future trajectories. Thanks to \mname, the trajectory decoder can attend to features of any agent at any previous timestep when inferring an agent's future position. To improve the diversity of sampled trajectories and avoid similar samples caused by random sampling, we further adopt a multi-agent trajectory sampler that can generate diverse and plausible multi-agent trajectories by mapping context to various configurations of all agents' latent codes.

We evaluate our method on well-established pedestrian datasets, ETH~\cite{pellegrini2009you} and UCY~\cite{lerner2007crowds}, and an autonomous driving dataset, nuScenes~\cite{caesar2020nuscenes}. On ETH/UCY and nuScenes, we outperform state-of-the-art multi-agent prediction methods with substantial performance improvement. We further conduct extensive ablation studies to show the superiority of \mname\ over various combinations of social and temporal models. We also demonstrate the efficacy of agent-aware attention against agent encoding.

To summarize, the main contributions of this paper are:
(1) We propose a new Transformer that simultaneously models the time and social dimensions of multi-agent trajectories with a sequence representation. (2) We propose a novel agent-aware attention mechanism that preserves the agent identity of each element in the multi-agent trajectory sequence. (3) We present a multi-agent forecasting framework that models the latent intent of all agents jointly to produce socially-plausible future trajectories. (4) Our approach substantially improves the state of the art on well-established pedestrian and autonomous driving datasets.

\section{Related Work}

\noindent\textbf{Sequence Modeling.} Sequences are an important representation of data such as video, audio, price, \etc Historically, RNNs (\eg, LSTMs~\cite{hochreiter1997long}, GRUs~\cite{chung2014empirical}) have achieved remarkable success in sequence modeling, with applications to speech recognition~\cite{xiong2018microsoft,miao2015eesen}, image captioning~\cite{xu2015show}, machine translation~\cite{luong2015effective}, human pose estimation~\cite{yuan20183d,kocabas2020vibe}, \etc In particular, RNNs have been the preferred temporal models for trajectory and motion forecasting. Many RNN-based methods model the trajectory pattern of pedestrians to predict their 2D future locations~\cite{alahi2016social,ivanovic2019trajectron,zhang2019sr}. Prior work has also used RNNs to model the temporal dynamics of 3D human pose~\cite{fragkiadaki2015recurrent,yuan2019ego,yuan2020residual}. With the invention of Transformers and positional encoding~\cite{vaswani2017attention}, many works start to adopt Transformers for sequence modeling due to their strong ability to capture long-range dependencies. Transformers have first dominated the natural language processing (NLP) domain across various tasks~\cite{devlin2018bert,lan2019albert,yang2019xlnet}. Beyond NLP, numerous visual Transformers have been proposed to tackle vision tasks, such as image classification~\cite{dosovitskiy2020image}, object detection~\cite{carion2020end}, and instance segmentation~\cite{wang2020end}. Recently, Transformers have also been used for trajectory forecasting. Transformer-TF~\cite{giuliari2020transformer} applies the standard Transformer to predict the future trajectories of each agent independently. STAR~\cite{yu2020spatio} uses separate temporal and spatial Transformers to forecast multi-agent trajectories. Interaction Transformer~\cite{li2020end} combines RNNs and Transformers for multi-agent trajectory modeling.
Different from prior work, Our \mname\ leverages a sequence representation of multi-agent trajectories and a novel agent-aware attention mechanism to preserve time and agent information in the sequence.

\vspace{1mm}
\noindent\textbf{Trajectory Prediction.}
Early work on trajectory prediction adopts a deterministic approach using models such as social forces~\cite{helbing1995social}, Gaussian process (GP)~\cite{wang2007gaussian}, and RNNs~\cite{alahi2016social,morton2016analysis,vemula2018social}. A thorough review of these deterministic methods is provided in~\cite{rudenko2020human}. As the future trajectory of an agent is uncertain and often multi-modal, recent trajectory prediction methods start to model the trajectory distribution with deep generative models~\cite{kingma2013auto,goodfellow2014generative,rezende2015variational} such as conditional variational autoencoders (CVAEs)~\cite{lee2017desire,yuan2019diverse,ivanovic2019trajectron,tang2019multiple,weng2020joint,salzmann2020trajectron++}, generative adversarial networks (GANs)~\cite{gupta2018social,sadeghian2019sophie,kosaraju2019social,zhao2019multi}, and normalizing flows (NFs)~\cite{rhinehart2018r2p2,rhinehart2019precog,guan2020generative}. Most of these methods follow a seq2seq structure~\cite{bahdanau2014neural,cho2014learning} and predict future trajectories using intermediate features of past trajectories. In contrast, our \mname-based trajectory prediction framework can directly attend to features of any agent at any previous timestep when inferring an agent's future position. Moreover, our approach models the future trajectories of all agents jointly to predict socially-aware trajectories.

\vspace{1mm}
\noindent\textbf{Social Interaction Modeling.} Methods for social interaction modeling can be categorized based on how they model the time and social dimensions. While RNNs~\cite{hochreiter1997long,chung2014empirical} and Transformers~\cite{vaswani2017attention} are the prefered temporal models~\cite{huang2019stgat,alahi2016social,yu2020spatio}, graph neural networks (GNNs)~\cite{kipf2016semi,li2015gated} are often employed as the social models for interaction modeling~\cite{kipf2018neural,li2020evolvegraph,kosaraju2019social}. One popular type of methods~\cite{kosaraju2019social,alahi2016social,gupta2018social} first uses temporal models to summarize trajectory features over time for each agent independently and then feeds the temporal features to social models to obtain socially-aware agent features. Alternatively, approaches like~\cite{salzmann2020trajectron++,huang2019stgat} first use social models to produce social features of each agent at each independent timestep and then apply temporal models to summarize the social features over time for each agent. One common characteristic of these prior works is that they model the time and social dimensions on separate levels. This can be suboptimal since it prevents an agent's feature at one time from directly interacting with another agent's feature at a different time, thus limiting the model's ability to capture long-range dependencies. Instead, our method models both the time and social dimensions simultaneously, allowing direct feature interaction across time and agents.

\section{Approach}
\label{sec:approach}
We formulate multi-agent trajectory prediction as modeling the generative future trajectory distribution of $N$ (variable) agents conditioned on their past trajectories. For observed timesteps $t \leq 0$, we represent the joint state of all $N$ agents at time $t$ as $\mb{X}^t = (\mb{x}_1^t, \mb{x}_2^t, \ldots, \mb{x}_N^t)$, where $\mb{x}_n^t \in \mathbb{R}^{d_s}$ is the state of agent $n$ at time $t$, which includes the position, velocity and (optional) heading angle of the agent. We denote the history of all agents as $\mb{X} = \left(\mb{X}^{-H}, \mb{X}^{-H+1}, \ldots, \mb{X}^{0}\right)$ which includes the joint agent state at all $H+1$ observed timesteps. Similarly, the joint state of all $N$ agents at future time $t$ ($t > 0$) is denoted as $\mb{Y}^t = (\mb{y}_1^t, \mb{y}_2^t, \ldots, \mb{y}_N^t)$, where $\mb{y}_n^t \in \mathbb{R}^{d_p}$ is the future position of agent $n$ at time $t$. We denote the future trajectories of all $N$ agents over $T$ future timesteps as $\mb{Y} = \left(\mb{Y}^{1}, \mb{Y}^{2}, \ldots, \mb{Y}^{T}\right)$. Depending on the data, optional contextual information $\mb{I}$ may also be given, such as a semantic map around the agents (annotations of sidewalks, road boundaries, \etc). Our goal is to learn a generative model $p_\theta(\mb{Y}|\mb{X}, \mb{I})$ where $\theta$ are the model parameters.

In the following, we first introduce the proposed agent-aware Transformer, \mname, for joint modeling of socio-temporal relations. We then present a stochastic multi-agent trajectory prediction framework that jointly models the latent intent of all agents. %

\subsection{\mname: Agent-Aware Transformers}
\label{sec:agent_former}
Our agent-aware Transformer, \mname, is a model that learns representations from multi-agent trajectories over both time and social dimensions simultaneously, in contrast to standard approaches that model the two dimensions in separate stages. \mname\ has two types of modules -- encoders and decoders, which follow the encoder and decoder design of the original Transformer~\cite{vaswani2017attention} but with two major differences: (1) it replaces positional encoding with a time encoder; (2) it uses a novel agent-aware attention mechanism instead of the scaled dot-product attention. As we will discuss below, these two modifications are motivated by a sequence representation of multi-agent trajectories that is suitable for Transformers.

\vspace{2mm}
\noindent\textbf{Multi-Agent Trajectories as a Sequence.}
The past multi-agent trajectories $\mb{X}$ can be denoted as a sequence $\mb{X} = \left(\mb{x}_1^{-H}, \ldots, \mb{x}_N^{-H}, \mb{x}_1^{-H+1}, \ldots, \mb{x}_N^{-H+1}, \ldots, \mb{x}_1^{0}, \ldots,  \mb{x}_N^{0}\right)$ of length $L_p=N\times(H + 1)$. Similarly, the future multi-agent trajectories can also be represented as a sequence $\mb{Y} = \left(\mb{y}_1^{1}, \ldots, \mb{y}_N^{1}, \mb{y}_1^{2}, \ldots, \mb{y}_N^{2}, \ldots, \mb{y}_1^{T}, \ldots,  \mb{y}_N^{T}\right)$ of length $L_f = N\times T$. We adopt this sequence representation to be compatible with Transformers. At first glance, it may seem that we can directly apply standard Transformers to these sequences to model temporal and social relations. However, there are \emph{two problems} with this approach:
(1)~\textbf{loss of \emph{time} information}, as Transformers have no notion of time when computing attention for each element (\eg, $\mb{x}_n^{t}$) \emph{w.r.t.} other elements in the sequence; for instance, $\mb{x}_n^{t}$ does not know $\mb{x}_{m}^{t}$ is a feature of the same timestep while $\mb{x}_n^{t+1}$ is a feature of the next timestep;
(2)~\textbf{loss of \emph{agent} information}, since Transformers do not consider agent identities when applying attention to each element, and elements of the same agent are not distinguished from elements of other agents; for example, when computing attention for $\mb{x}_n^{t}$, both $\mb{x}_n^{t+1}$ and $\mb{x}_{m}^{t+1}$ are treated the same, disregarding the fact that $\mb{x}_n^{t+1}$ is from the same agent while $\mb{x}_{m}^{t+1}$ is from a different agent. Below, we present the solutions to these two problems -- (1)~time encoder and (2) agent-aware attention.

\vspace{2mm}
\noindent\textbf{Time Encoder.} To inform \mname\ about the timestep associated with each element in the trajectory sequence, we employ a time encoder similar to the positional encoding in the original Transformer. Instead of encoding the position of each element based on its index in the sequence, we compute a timestamp feature based on the timestep $t$ of the element. The timestamp uses the same sinusoidal design as the positional encoding. Let us take the past trajectory sequence $\mb{X}$ as an example. For each element $\mathbf{x}_n^t$, the timestamp feature $\bs{\tau}_n^t \in \mathbb{R}^{d_\tau}$ is defined as
\begin{equation*}
\bs{\tau}_n^t(k) =
\begin{cases}
    \sin((t + H)/10000^{k/d_\tau}),              & k\text{ is even}\\
    \cos((t + H)/10000^{(k-1)/d_\tau}),          & k\text{ is odd}
\end{cases}
\end{equation*}
where $\bs{\tau}_n^t(k)$ denotes the $k$-th feature of $\bs{\tau}_n^t$ and $d_\tau$ is the feature dimension of the timestamp. The time encoder outputs a timestamped sequence $\bar{\mb{X}}$ and each element $\bar{\mb{x}}_n^t \in \mathbb{R}^{d_\tau}$ in $\bar{\mb{X}}$ is computed as $ \bar{\mb{x}}_n^t = \mb{W}_2(\mb{W}_1\mb{x}_n^t\oplus \bs{\tau}_n^t)$ where $\mb{W}_1 \in \mathbb{R}^{d_\tau\times d_s}$ and $\mb{W}_2 \in \mathbb{R}^{d_\tau\times 2d_\tau}$ are weight matrices and $\oplus$ denotes concatenation.

\vspace{2mm}
\noindent\textbf{Agent-Aware Attention.} To preserve agent information in the trajectory sequence, it may be tempting to employ a similar strategy to the time encoder, such as an agent encoder that assigns an agent index-based encoding to each element in the sequence. However, using such agent encoding is not effective as we will show in the experiments. The reason is that, different from time which is naturally ordered, there is no innate ordering between agents, and assigning encodings based on agent indices will break the required permutation invariance of agents and create artificial dependencies on agent indices in the model.

We tackle the loss of agent information from a different angle by proposing a novel agent-aware attention mechanism. The agent-aware attention takes as input keys $\mb{K}$, queries $\mb{Q}$ and values $\mb{V}$, each of which uses the sequence representation of multi-agent trajectories. As an example, let the keys $\mb{K}$ and values $\mb{V}$ be the past trajectory sequence $\mb{X}\in \mathbb{R}^{L_p \times d_s}$, and let the queries $\mb{Q}$ be the future trajectory sequence $\mb{Y}\in \mathbb{R}^{L_f \times d_p}$. Recall that $\mb{X}$ is of length $L_p = N\times(H+1)$ as $\mb{X}$ contains the trajectory features of $N$ agents of $H+1$ past timesteps; $\mb{Y}$ is of length $L_f = N\times T$ containing trajectory features of $T$ future timesteps. The output of agent-aware attention is computed as

\vspace{-3mm}
\begin{small}
\begin{align}
&\text{AgentAwareAttention}(\mb{Q}, \mb{K}, \mb{V}) = \text{softmax}\left(\frac{\mb{A}}{\sqrt{d_k}}\right)V \\
\label{eq:attn}
&\mb{A} = \mb{M} \odot (\mb{Q}_{self} \mb{K}_{self}^T) + (1 - \mb{M}) \odot (\mb{Q}_{other}\mb{K}_{other}^T) \\
&\mb{Q}_{self} = \mb{Q}\mb{W}_{self}^Q, \quad\;\; \mb{K}_{self} = \mb{K}\mb{W}_{self}^K\\
&\mb{Q}_{other} = \mb{Q}\mb{W}_{other}^Q, \quad \mb{K}_{other} = \mb{K}\mb{W}_{other}^K
\end{align}
\vspace{-3mm}
\end{small}

\begin{figure}
    \centering
    \includegraphics[width=\linewidth]{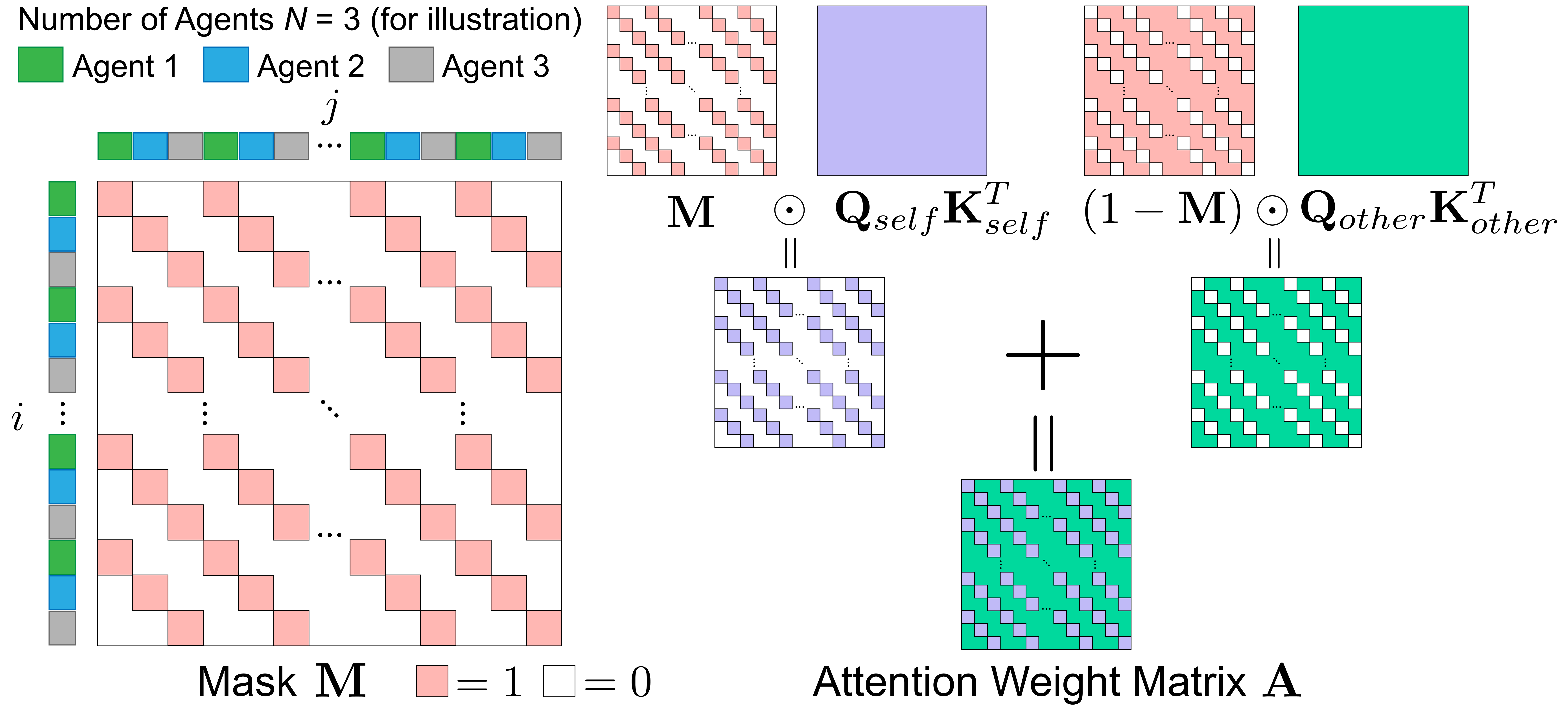}
    \caption{\textbf{Illustration of agent-aware attention.} The mask $\mb{M}$ allows the attention weights in $\mb{A}$ to be computed differently based on whether the $i$-th query and $j$-th key belong to the same agent. }
    \label{fig:attention}
    \vspace{-2mm}
\end{figure}

\begin{figure*}[ht]
    \centering
    \includegraphics[width=\textwidth]{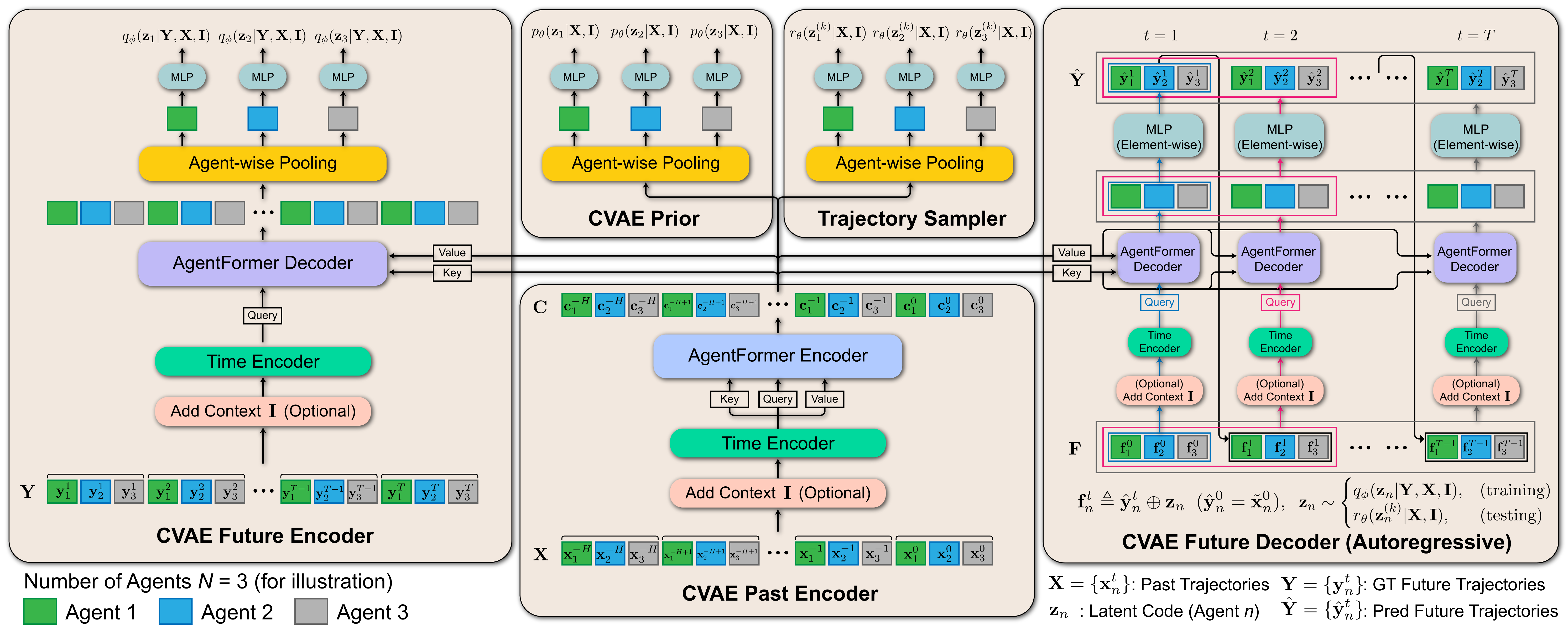}
    \caption{\textbf{Overview of our \mname-based multi-agent trajectory prediction framework.}}
    \label{fig:overview}
    \vspace{-2mm}
\end{figure*}

\noindent where $\odot$ denotes element-wise product and we use two sets of projections $\{\mb{W}_{self}^Q,\mb{W}_{self}^K\}$ and $\{\mb{W}_{other}^Q,\mb{W}_{other}^K\}$ to generate projected keys $\mb{K}_{self},\mb{K}_{other}\in \mathbb{R}^{L_p\times d_k}$ and queries $\mb{Q}_{self},\mb{Q}_{other}\in \mathbb{R}^{L_f\times d_k}$ with key (query) dimension $d_k$. Each element $A_{ij}$ in the attention weight matrix $\mb{A}$ represents the attention weight between the $i$-th query $\mb{q}_i$ and \mbox{$j$-th} key $\mb{k}_j$. As illustrated in Fig.~\ref{fig:attention}, when computing the attention weight matrix $\mb{A} \in \mathbb{R}^{L_f\times L_p}$, we also use a mask $\mb{M} \in \mathbb{R}^{L_f\times L_p}$ which is defined as
\begin{equation}
M_{ij} = \mathbbm{1}  (i \text{ mod } N = j \text{ mod } N)
\end{equation}
where $M_{ij}$ denotes each element inside the mask $\mb{M}$ and $\mathbbm{1}(\cdot)$ denotes the indicator function. As $\cdot$ mod $N$ computes the agent index of a query/key, $M_{ij}$ equals to one if the $i$-th query $\mb{q}_i$ and $j$-th key $\mb{k}_j$ belongs to the same agent, and $M_{ij}$ equals to zero otherwise, as shown in Fig.~\ref{fig:attention}. Using the mask $\mb{M}$, Eq.~\eqref{eq:attn} computes each element $A_{ij}$ of the attention weight matrix $\mb{A}$ differently based on the agreement of agent identity: If $\mb{q}_i$ and $\mb{k}_j$ have the same agent identity, $A_{ij}$ is computed using the projected queries $\mb{Q}_{self}$ and keys $\mb{K}_{self}$ designated for intra-agent attention (agent to itself); If $\mb{q}_i$ and $\mb{k}_j$ have different agent identities, $A_{ij}$ is computed using the projected queries $\mb{Q}_{other}$ and keys $\mb{K}_{other}$ designated for inter-agent attention (agent to other agents). In this way, the agent-aware attention learns to attend to elements of the same agent in the sequence differently than elements of other agents, thus preserving the notion of agent identity. Note that \mname\ only uses agent-aware attention to replace the scaled dot-product attention in the original Transformer and still allows multi-head attention to learn distributed representations.

\vspace{2mm}
\noindent\textbf{Encoding Agent Connectivity.}
\mname\ can also encode rule-based agent connectivity information by masking out the attention weights between unconnected agents. Specifically, we define that two agents $n$ and $m$ are connected if their distance $D_{nm}$ at the current time ($t=0$) is smaller than a threshold $\eta$. If agents $n$ and $m$ are not connected, we set the attention weight $A_{ij} = -\infty$ between any query $\mb{q}_i$ of agent $n$ and any key $\mb{k}_j$ of agent $m$.

\subsection{Multi-Agent Prediction with \mname}
\label{sec:cvae}
Having introduced \mname\ for modeling temporal and social relations, we are now ready to apply it in our multi-agent trajectory prediction framework based on CVAEs. As discussed at the start of Sec.~\ref{sec:approach}, the goal of multi-agent trajectory prediction is to model the future trajectory distribution $p_\theta(\mb{Y}|\mb{X},\mb{I})$ conditioned on past trajectories $\mb{X}$ and contextual information $\mb{I}$. To account for stochasticity and multi-modality in each agent's future behavior, we introduce latent variables $\mb{Z} = \{\mb{z}_1, \ldots, \mb{z}_N\}$ where $\mb{z}_n \in \mathbb{R}^{d_z}$ represents the latent intent of agent $n$. We can then rewrite the future trajectory distribution as
\vspace{-1mm}
\begin{equation}
\label{eq:cvae}
p_\theta(\mb{Y}|\mb{X},\mb{I}) = \int p_\theta(\mb{Y}|\mb{Z},\mb{X},\mb{I}) p_\theta(\mb{Z}|\mb{X},\mb{I}) d\mb{Z}\,,
\vspace{-1mm}
\end{equation}
where $p_\theta(\mb{Z}|\mb{X},\mb{I}) = \prod_{n=1}^N p_\theta(\mb{z}_n|\mb{X},\mb{I})$ is a conditional Gaussian prior factorized over agents and $p_\theta(\mb{Y}|\mb{Z},\mb{X},\mb{I})$ is a conditional likelihood model. To tackle the intractable integral in Eq.~\eqref{eq:cvae}, we use the negative evidence lower bound (ELBO) $\mathcal{L}_{elbo}$ in the CVAE as our loss function:
\begin{equation}
\label{eq:elbo}
\begin{aligned}
\mathcal{L}_{elbo} = & -\mathbb{E}_{q_\phi(\mb{Z}|\mb{Y},\mb{X},\mb{I})}[\log p_\theta(\mb{Y}|\mb{Z},\mb{X},\mb{I})] \\
& + \text{KL}(q_\phi(\mb{Z}|\mb{Y},\mb{X},\mb{I})\| p_\theta(\mb{Z}|\mb{X},\mb{I}))\,,
\end{aligned}
\end{equation}
where $q_\phi(\mb{Z}|\mb{Y},\mb{X},\mb{I}) = \prod_{n=1}^N q_\phi(\mb{z}_n|\mb{Y},\mb{X},\mb{I})$ is an approximate posterior distribution factorized over agents and parametrized by $\phi$. In our probabilistic formulation, the latent codes $\mb{Z}$ of all agents in the posterior $q_\phi(\mb{Z}|\mb{Y},\mb{X},\mb{I})$ are jointly inferred from the future trajectories $\mb{Y}$ of all agents; similarly, the future trajectories $\mb{Y}$ in the conditional likelihood $p_\theta(\mb{Y}|\mb{Z},\mb{X},\mb{I})$ are modeled using the latent codes~$\mb{Z}$ of all agents. This design allows each agent's latent intent represented by $\mb{z}_n$ to affect not just its own future trajectory but also the future trajectories of other agents, which enables us to generate socially-aware multi-agent trajectories. Having described the probabilistic formulation, we now introduce the detailed model architecture as outlined in Fig.~\ref{fig:overview}.

\vspace{2mm}
\noindent\textbf{Encoding Context (Semantic Map).}
As aforementioned, our model can optionally take as input contextual information $\mb{I}$ if provided by the data. Here, we assume $\mb{I}\in \mathbb{R}^{H_0\times W_0\times C}$ is a semantic map around the agents at the current timestep ($t=0$) with annotated semantic information (\eg, sidewalks, crosswalks, and road boundaries). For each agent $n$, we rotate $\mb{I}$ to align with the agent's heading angle and crop an image patch $\mb{I}_n \in \mathbb{R}^{H\times W\times C}$ around the agent. We use a hand-designed convolutional neural network (CNN) to extract visual features $\mb{v}_n$ from $\mb{I}_n$, which will later be used by other modules in the model.

\vspace{2mm}
\noindent\textbf{CVAE Past Encoder.}
The past encoder starts with the multi-agent past trajectory sequence $\mb{X}$. If the semantic map $\mb{I}$ is provided, the past encoder concatenates each element $\mb{x}_n^t \in \mb{X}$ with the corresponding visual feature $\mb{v}_n$ of agent $n$. The new sequence is then fed into the time encoder to obtain a timestamped sequence, which is then input to the \mname\ encoder as keys, queries, and values. The output of the encoder is a past feature sequence $\mb{C} =  \left(\mb{c}_1^{-H}, \ldots \mb{c}_N^{-H}, \mb{c}_1^{-H+1}, \ldots \mb{c}_N^{-H+1}, \ldots, \mb{c}_1^{0}, \ldots,  \mb{c}_N^{0}\right)$ that summarizes the past agent trajectories $\mb{X}$ and context $\mb{I}$.

\vspace{2mm}
\noindent\textbf{CVAE Prior.}
The prior module first performs an agent-wise pooling that computes a mean agent feature $\mb{C}_n$ from the past features across timesteps: $\mb{C}_n = \text{mean}(\mb{c}_n^{-H}, \ldots, \mb{c}_n^{0})$. We then use a multilayer perceptron (MLP) to map $\mb{C}_n$ to the Gaussian parameters $(\bs{\mu}_n^p, \bs{\sigma}_n^p)$ of the prior distribution $p_\theta(\mb{z}_n|\mb{X},\mb{I}) = \mathcal{N}(\bs{\mu}_n^p,\text{Diag}( \bs{\sigma}_n^p)^2)$.

\vspace{2mm}
\noindent\textbf{CVAE Future Encoder.}
Given the multi-agent future trajectory sequence $\mb{Y}$, similar to the past encoder, the future encoder appends visual features from the semantic map $\mb{I}$ to $\mb{Y}$ and feeds the resulting sequence to the time encoder to produce a timestamped sequence. The timestamped sequence is then input as queries to the \mname\ decoder along with the past feature sequence $\mb{C}$ which serves as both keys and values. We use the \mname\ decoder here because it allows the feature extraction of $\mb{Y}$ to condition on $\mb{X}$ through $\mb{C}$, thus effectively modeling the $\mb{X}$-conditioning in the posterior $q_\phi(\mb{Z}|\mb{Y},\mb{X},\mb{I})$. We then perform an agent-wise mean pooling across timesteps on the output sequence of the \mname\ decoder to extract a feature for each agent. Each agent feature is then input to an MLP to obtain the Gaussian parameters $(\bs{\mu}_n^q, \bs{\sigma}_n^q)$ of the approximate posterior distribution $q_\phi(\mb{z}_n|\mb{Y},\mb{X},\mb{I}) = \mathcal{N}(\bs{\mu}_n^q,\text{Diag}( \bs{\sigma}_n^q)^2)$.

\vspace{2mm}
\noindent\textbf{CVAE Future Decoder.}
Unlike the original Transformer decoder, our future trajectory decoder is autoregressive, which means it outputs trajectories one step at a time and feeds the currently generated trajectories back into the model to produce the trajectories of the next timestep. This design mitigates compounding errors during test time at the expense of training speed. Starting from an initial sequence $(\hat{\mb{y}}_1^0,\ldots,\hat{\mb{y}}_N^0)$ where $\hat{\mb{y}}_n^0 = \tilde{\mb{x}}_n^0$ ($\tilde{\mb{x}}_n^0$ is the position feature inside $\mb{x}_n^0$), the future decoder module maps an input sequence $(\hat{\mb{y}}_1^0,\ldots,\hat{\mb{y}}_N^0, \ldots, \hat{\mb{y}}_1^{t'},\ldots,\hat{\mb{y}}_N^{t'})$ to an output sequence $(\hat{\mb{y}}_1^1,\ldots,\hat{\mb{y}}_N^1, \ldots, \hat{\mb{y}}_1^{t'+1},\ldots,\hat{\mb{y}}_N^{t'+1})$ and grows the input sequence into $(\hat{\mb{y}}_1^0,\ldots,\hat{\mb{y}}_N^0, \ldots, \hat{\mb{y}}_1^{t'+1},\ldots,\hat{\mb{y}}_N^{t'+1})$. By autoregressively applying the decoder $T$ times, we obtain the output sequence $\hat{\mb{Y}} = (\hat{\mb{y}}_1^1,\ldots,\hat{\mb{y}}_N^1, \ldots, \hat{\mb{y}}_1^{T},\ldots,\hat{\mb{y}}_N^{T}) $. Inside the future decoder module (Fig.~\ref{fig:overview}\,(Right)), we first form a feature sequence $\mb{F} = (\mb{f}_1^0,\ldots,\mb{f}_N^0, \ldots,\mb{f}_1^{t'},\ldots,\mb{f}_N^{t'})$ where $\mb{f}_n^{t} = \hat{\mb{y}}_n^{t} \oplus \mb{z}_n$, thus concatenating the currently generated trajectories with the corresponding latent codes. The latent codes are sampled from the approximate posterior during training but from the trajectory sampler (as discussed below) at test time. The feature sequence $\mb{F}$ is then concatenated with the semantic map features and timestamped before being input as queries to the \mname\ decoder alongside the past feature sequence $\mb{C}$ which serves as keys and values. The \mname\ decoder enables the future trajectories to directly attend to features of any agent at any previous timestep (\eg, $\mb{c}_3^{-H}$ or $\hat{\mb{y}}_2^{1}$), allowing the model to effectively infer future trajectories based on the whole agent history. We use proper masking inside the \mname\ decoder to enforce causality of the decoder output sequence. Each element of the output sequence is then passed through an MLP to generate the decoded future agent position $\hat{\mb{y}}_n^{t}$. As we use a Gaussian to model the conditional likelihood $p_\theta(\mb{Y}|\mb{Z},\mb{X},\mb{I}) = \mathcal{N}(\hat{\mb{Y}},I/\beta)$, where $I$ is the identity matrix and $\beta$ is a weighting factor, the first term in Eq.~\eqref{eq:elbo} equals the mean squred error (MSE): $\mathcal{L}_{mse} = \frac{1}{2\beta}\|\mb{Y} - \hat{\mb{Y}}\|^2$.

\vspace{2mm}
\noindent\textbf{Trajectory Sampler.}
We adapt a diversity sampling technique, DLow~\cite{yuan2020dlow}, to our multi-agent trajectory prediction setting and employ a trajectory sampler to produce diverse and plausible trajectories once our CVAE model is trained. The trajectory sampler generates $K$ sets of latent codes $\{\mb{Z}^{(1)}, \ldots, \mb{Z}^{(K)}\}$ where each set $\mb{Z}^{(k)} = \{\mb{z}_1^{(k)}, \ldots, \mb{z}_N^{(k)}\}$ contains the latent codes of all agents and can be decoded by the CVAE decoder into a multi-agent future trajectory sample $\hat{\mb{Y}}^{(k)}$. Each latent code $\mb{z}_n^{(k)} \in \mb{Z}^{(k)}$ is generated by a linear transformation of a Gaussian noise $\bs{\epsilon}_n \in \mathbb{R}^{d_z}$:
\begin{equation}
\label{eq:samp}
\mb{z}_n^{(k)} = \mb{A}_n^{(k)}\bs{\epsilon}_n + \mb{b}_n^{(k)}, \quad \bs{\epsilon}_n \sim \mathcal{N}(\mb{0}, I),
\end{equation}
where $\mb{A}_n^{(k)} \in \mathbb{R}^{d_z \times d_z}$ is a non-singular matrix and $\mb{b}_n^{(k)} \in \mathbb{R}^{d_z}$ is a vector. Eq.~\eqref{eq:samp} induces a Gaussian sampling distribution $r_\theta(\mb{z}_n^{(k)}|\mb{X}, \mb{I})$ over $\mb{z}_n^{(k)}$. The distribution is conditioned on $\mb{X}$ and $\mb{I}$ because its inner parameters $\{\mb{A}_n^{(k)},\mb{b}_n^{(k)}\}$ are generated by the trajectory sampler module (Fig.~\ref{fig:overview}) through agent-wise pooling of the past feature sequence $\mb{C}$ and an MLP. The trajectory sampler loss is defined as

\vspace{-3mm}
\begin{small}
\begin{equation}
\begin{aligned}
\label{eq:samp_loss}
&\mathcal{L}_{samp} = \min_k \|\hat{\mb{Y}}^{(k)} - \mb{Y}\|^2\\
& + \sum_{n=1}^N\text{KL}(r_\theta(\mb{z}_n^{(k)}|\mb{X}, \mb{I})\| p_\theta(\mb{z}_n|\mb{X}, \mb{I})) \\
& + \frac{1}{K(K-1)} \sum_{k_1=1}^{K} \sum_{k_1 \neq k_2}^{K} \exp \left(-\frac{\|\hat{\mb{Y}}^{(k_1)} - \hat{\mb{Y}}^{(k_2)}\|^2}{\sigma_{d}}\right),
\end{aligned}
\end{equation}
\vspace{-2mm}
\end{small}

\noindent where $\sigma_d$ is a scaling factor. The first term encourages the future trajectory samples $\hat{\mb{Y}}^{(k)}$ to cover the ground truth $\mb{Y}$. The second KL term encourages each latent code $\mb{z}_n^{(k)}$ to follow the prior and be plausible; the KL can be computed analytically as both distributions inside are Gaussians. The third term encourages diversity among the future trajectory samples $\hat{\mb{Y}}^{(k)}$ by penalizing small pairwise distance. When training the trajectory sampler with Eq.~\eqref{eq:samp_loss}, we freeze the weights of the CVAE modules. At test time, we sample latent codes $\{\mb{Z}^{(1)}, \ldots, \mb{Z}^{(K)}\}$ using the trajectory sampler instead of sampling from the CVAE prior and decode the latent codes into trajectory samples $\{\hat{\mb{Y}}^{(1)}, \ldots, \hat{\mb{Y}}^{(K)}\}$.

\section{Experiments}
\label{sec:exp}

\noindent\textbf{Datasets.}
We evaluate our method on well-established public datasets: the ETH~\cite{pellegrini2009you}, UCY~\cite{lerner2007crowds}, and nuScenes~\cite{caesar2020nuscenes} datasets. The ETH/UCY datasets are the major benchmark for pedestrian trajectory prediction. There are five datasets in ETH/UCY, each of which contains pedestrian trajectories captured at 2.5Hz in multi-agent social scenarios with rich interaction. nuScenes is a recent large-scale autonomous driving dataset, which consists of 1000 driving scenes with each scene annotated at 2Hz. nuScenes also provides HD semantic maps with 11 semantic classes.

\vspace{2mm}
\noindent\textbf{Metrics.}
We report the minimum average displacement error $\text{ADE}_K$ and final displacement error $\text{FDE}_K$ of $K$ trajectory samples of each agent compared to the ground truth: $\text{ADE}_K = \frac{1}{T}\min_{k=1}^{K}\sum_{t=1}^T\|\hat{\mb{y}}_n^{t,(k)} - \mb{y}_n^{t}\|^2, \quad \text{FDE}_K = \min_{k=1}^{K}\|\hat{\mb{y}}_n^{T,(k)} - \mb{y}_n^{T}\|^2$, where  $\hat{\mb{y}}_n^{t,(k)}$ denotes the future position of agent $n$ at time $t$ in the $k$-th sample and $\mb{y}_n^{T}$ is the corresponding ground truth. $\text{ADE}_K$ and $\text{FDE}_K$ are the standard metrics for trajectory prediction~\cite{gupta2018social,sadeghian2019sophie,salzmann2020trajectron++,phan2020covernet,chai2020multipath}.

\vspace{2mm}
\noindent\textbf{Evaluation Protocol.}
For the ETH/UCY datasets, we adopt a leave-one-out strategy for evaluation, following prior work~\cite{gupta2018social,sadeghian2019sophie,salzmann2020trajectron++,mangalam2020not,yu2020spatio}. We forecast 2D future trajectories of 12 timesteps (4.8s) based on observed trajectories of 8 timesteps (3.2s). Similar to most prior works, we do not use any semantic/visual information for ETH/UCY for fair comparisons. All metrics are computed with $K=20$ samples. For the nuScenes dataset, following prior work~\cite{phan2020covernet,chai2020multipath,cui2019multimodal,ma2020diverse}, we use the vehicle-only train-val-test split provided by the nuScenes prediction challenge and predict 2D future trajectories of 12 timesteps (6s) based on observed trajectories of 4 timesteps (2s). We report results with metrics computed using $K=1, 5 \text{ and } 10$ samples.

\vspace{2mm}
\noindent\textbf{Implementation Details.}
For all datasets, we represent trajectories in a scene-centered coordinate where the origin is the mean position of all agents at $t=0$. The future decoder in Fig.~\ref{fig:overview} outputs the offset to the agent's current position $\tilde{\mb{x}}_n^0$, so $\tilde{\mb{x}}_n^0$ is added to obtain $\hat{\mb{y}}_n^t$ for each element in the output sequence. Following prior work~\cite{salzmann2020trajectron++,yu2020spatio}, random rotation of the scene is adopted for data augment.
Our multi-agent prediction model (Fig.~\ref{fig:overview}) uses two stacks (defined in \cite{vaswani2017attention}) of identical layers in each \mname\ encoder/decoder with 0.1 dropout rate. The dimensions $d_k,d_v,d_\tau$ of keys, queries, and timestamps in \mname\ are all set to 256, and the hidden dimension of feedforward layers is 512. The number of heads for multi-head agent-aware attention is 8. All MLPs in the model have hidden dimensions (512, 256). For the CVAE, the latent code dimension $d_z$ is 32, the coefficient $\beta$ of the MSE loss equals 1, and we clip the maximum value of the KL term in $L_{elbo}$ (Eq.~\eqref{eq:elbo}) down to 2. We also use the variety loss in SGAN~\cite{gupta2018social} in addition to $L_{elbo}$. The agent connectivity threshold $\eta$ is set to 100. We train the CVAE model using the Adam optimizer~\cite{kingma2014adam} for 100 epochs on ETH/UCY and nuScenes. We use an initial learning rate of $10^{-4}$ and halve the learning rate every 10 epochs. More details including the CNN for encoding semantic maps and the training procedure of the trajectory sampler can be found in Appendix~\ref{sec:supp_details}.

\setlength{\tabcolsep}{3pt}
\begin{table}[t]
\footnotesize
\centering
\resizebox{\linewidth}{!}{
\begin{tabular}{@{\hskip 1mm}l@{\hskip 1mm}|ccccc|@{\hskip 1mm}c@{\hskip 1mm}}
\toprule
\multicolumn{1}{c|}{\multirow{3}{*}[2pt]{Method}} & \multicolumn{6}{c}{$\text{ADE}_{20}/\text{FDE}_{20} \downarrow$ (m), $K=20$ Samples} \\
\cmidrule(l{0.8mm}r{0.5mm}){2-7}
 & ETH & Hotel & Univ &  Zara1 & Zara2 & Average\\ \midrule%
SGAN~\cite{gupta2018social} & 0.81/1.52 & 0.72/1.61 & 0.60/1.26 & 0.34/0.69 & 0.42/0.84 & 0.58/1.18 \\
SoPhie~\cite{sadeghian2019sophie} & 0.70/1.43 & 0.76/1.67 & 0.54/1.24 & 0.30/0.63 & 0.38/0.78 & 0.54/1.15 \\
Transformer-TF~\cite{giuliari2020transformer} & 0.61/1.12 & 0.18/0.30 & 0.35/0.65 & 0.22/0.38 & 0.17/0.32 & 0.31/0.55 \\
STAR~\cite{yu2020spatio} & 0.36/0.65 & 0.17/0.36 & 0.31/0.62 & 0.26/0.55 & 0.22/0.46 & 0.26/0.53 \\
PECNet~\cite{mangalam2020not} & 0.54/0.87 & 0.18/0.24 & 0.35/0.60 & 0.22/0.39 & 0.17/0.30 & 0.29/0.48 \\
Trajectron++~\cite{salzmann2020trajectron++} & \textbf{0.39}/0.83 & \textbf{0.12}/\textbf{0.21} & \textbf{0.20}/\textbf{0.44} & \textbf{0.15}/0.33 & \textbf{0.11}/0.25 & \textbf{0.19}/0.41 \\
Ours (\mname) & 0.45/\textbf{0.75} & 0.14/0.22 & 0.25/0.45 & 0.18/\textbf{0.30} & 0.14/\textbf{0.24} & 0.23/\textbf{0.39} \\
\bottomrule
\end{tabular}
}
\vspace{-1mm}
\caption{\textbf{Baseline comparisons} on the ETH/UCY datasets.}
\label{table:eth}
\vspace{-2.5mm}
\end{table}
\setlength{\tabcolsep}{4pt}
\begin{table}[t]
\footnotesize
\centering
\resizebox{\linewidth}{!}{
\begin{tabular}{@{\hskip 1mm}l@{\hskip 1mm}|cccc}
\toprule
\multicolumn{1}{c|}{\multirow{3}{*}[2pt]{Method}} & \multicolumn{2}{c}{$K=5$ Samples} & \multicolumn{2}{c}{$K=10$ Samples} \\
\cmidrule(l{0.8mm}r{0.8mm}){2-3}
\cmidrule(l{0.8mm}r{0.8mm}){4-5}
 & $\text{ADE}_5\downarrow$ & $\text{FDE}_5\downarrow$ & $\text{ADE}_{10}\downarrow$ & $\text{FDE}_{10}\downarrow$ \\ \midrule
MTP~\cite{cui2019multimodal} & 2.93 & - & 2.93 & - \\
MultiPath~\cite{chai2020multipath} & 2.32 & - & 1.96 & - \\
CoverNet~\cite{phan2020covernet} & 1.96 & - & 1.48 & - \\
DSF-AF~\cite{ma2020diverse} & 2.06 & 4.67 & 1.66 & 3.71 \\
DLow-AF~\cite{yuan2020dlow} & 2.11 & 4.70 & 1.78 & 3.58 \\
Trajectron++~\cite{salzmann2020trajectron++} & 1.88 & - & 1.51 & - \\
Ours (\mname) & \textbf{1.86} & \textbf{3.89} & \textbf{1.45} & \textbf{2.86}\\
\bottomrule
\end{tabular}
}
\vspace{-.5mm}
\caption{\textbf{Baseline comparisons} on the nuScenes dataset.}
\label{table:nuscene}
\vspace{-3.5mm}
\end{table}

\subsection{Results}
\noindent\textbf{Baseline Comparisons.}
On the ETH/UCY datasets, we compare our approach with current state-of-the-art methods -- Trajectron++~\cite{salzmann2020trajectron++}, PECNet~\cite{mangalam2020not}, STAR~\cite{yu2020spatio}, and Transformer-TF~\cite{giuliari2020transformer} -- as well as common baselines -- SGAN~\cite{gupta2018social} and Sophie~\cite{sadeghian2019sophie}. The performance of all methods is summarized in Table~\ref{table:eth}, where we use officially-reported results for the baselines. We can observe that our \mname\ achieves very competitive performance and attains the best FDE. Particularly, our approach significantly outperforms prior Transformer-based methods, Transformer-TF~\cite{giuliari2020transformer} and STAR~\cite{yu2020spatio}. As FDE measures the final displacement error of predicted trajectories, it places more emphasis on a method's ability to predict distant futures than ADE. We believe the strong performance of our method in FDE can be attributed to the design of \mname, which can model long-range trajectory dependencies effectively by directly attending to features of any agent at any previous timestep when inferring an agent's future position.

Compared to ETH/UCY, the trajectories in nuScenes are much longer as we evaluate with a longer time horizon (6s) and vehicles are much faster than pedestrians. Thus, nuScenes presents a different challenge for multi-agent prediction methods. On the nuScenes dataset, we evaluate our approach against state-of-the-art vehicle prediction methods -- Trajectron++~\cite{salzmann2020trajectron++}, MTP~\cite{cui2019multimodal}, MultiPath~\cite{chai2020multipath}, CoverNet~\cite{phan2020covernet}, DSF-AF~\cite{ma2020diverse}, and DLow-AF~\cite{yuan2020dlow}. We report the performance of all methods in Table~\ref{table:nuscene}, where the results of Trajectron++ are taken from the nuScenes prediction challenge leaderboard, the performance of DLow-AF is from ~\cite{ma2020diverse}, and we also use the officially-reported results for the other baselines. The FDE of some baselines is not available since the number has not been reported. We can see that our approach, \mname, outperforms the baselines, especially the strong model Trajectron++~\cite{salzmann2020trajectron++}, consistently in ADE and FDE for both 5 and 10 sample settings.

\setlength{\tabcolsep}{3pt}
\begin{table}[t]
\footnotesize
\centering
\resizebox{\linewidth}{!}{
\begin{tabular}{@{\hskip 1mm}cc|ccccc|@{\hskip 1mm}c@{\hskip 1mm}}
\toprule
\multicolumn{2}{c|}{Model} & \multicolumn{6}{c}{$\text{ADE}_{20}/\text{FDE}_{20} \downarrow$ (m), $K=20$ Samples} \\
\cmidrule(l{0mm}r{0.5mm}){1-2}\cmidrule(l{0.5mm}r{1mm}){3-8}
 \hspace{1.5mm} Social \hspace{1.5mm} & Temporal & ETH & Hotel & Univ &  Zara1 & Zara2 & Average\\ \midrule%
GCN & LSTM & 0.57/0.90 & 0.20/0.34 & 0.29/0.52 & 0.24/0.44 & 0.23/0.42 & 0.31/0.52 \\
GCN & TF & 0.56/0.93 & 0.15/0.28 & 0.28/0.51 & 0.24/0.45 & 0.19/0.35 & 0.28/0.50 \\
TF & LSTM & 0.55/0.91 & 0.18/0.31 & 0.28/0.50 & 0.24/0.44 & 0.21/0.39 & 0.29/0.51 \\
TF & TF & 0.50/0.82 & 0.15/0.27 & 0.28/0.52 & 0.22/0.42 & 0.16/0.31 & 0.26/0.47 \\
\midrule
\multicolumn{2}{c|}{Joint Socio-Temporal} & ETH & Hotel & Univ &  Zara1 & Zara2 & Average\\ \midrule%
\multicolumn{2}{l|}{Ours w/o joint latent}  & 0.49/0.77 & 0.15/0.25 & 0.29/0.52 & 0.22/0.41 & 0.18/0.33 & 0.27/0.46 \\
\multicolumn{2}{l|}{Ours w/o AA attention} & 0.49/0.80 & 0.15/0.25 & 0.31/0.54 & 0.23/0.41 & 0.19/0.34 & 0.27/0.47 \\
\multicolumn{2}{l|}{Ours w/ agent encoding} & 0.48/0.78 & \textbf{0.14}/0.23 & 0.32/0.55 & 0.22/0.40 & 0.19/0.34 & 0.27/0.46 \\
\multicolumn{2}{l|}{Ours (\mname)} & \textbf{0.45}/\textbf{0.75} & \textbf{0.14}/\textbf{0.22} & \textbf{0.25}/\textbf{0.45} & \textbf{0.18}/\textbf{0.30} & \textbf{0.14}/\textbf{0.24} & \textbf{0.23}/\textbf{0.39} \\
\bottomrule
\end{tabular}
}
\vspace{-0.5mm}
\caption{\textbf{Ablation studies} on the ETH/UCY datasets. ``TF'' means Transformer and ``AA Attention'' denotes agent-aware attention.}
\label{table:eth_abl}
\vspace{-2mm}
\end{table}
\setlength{\tabcolsep}{4pt}
\begin{table}[t]
\footnotesize
\centering
\begin{tabular}{@{\hskip 1mm}cc|cccc}
\toprule
\multicolumn{2}{c|}{Model} & \multicolumn{2}{c}{$K=5$ Samples} & \multicolumn{2}{c}{$K=10$ Samples}\\
\cmidrule(l{0mm}r{0.5mm}){1-2}
\cmidrule(l{0.5mm}r{0.8mm}){3-4}
\cmidrule(l{0.8mm}r{0.8mm}){5-6}
 \hspace{1.5mm} Social \hspace{1.5mm} & Temporal & $\text{ADE}_5\downarrow$ & $\text{FDE}_5\downarrow$ & $\text{ADE}_{10}\downarrow$ & $\text{FDE}_{10}\downarrow$\\ \midrule%
GCN & LSTM & 2.17 & 4.42 & 1.57 & 3.09 \\
GCN & TF & 2.03 & 4.36 & 1.52 & 2.95 \\
TF & LSTM & 2.12 & 4.48 & 1.69 & 3.31 \\
TF & TF & 1.99 & 4.12 & 1.54 & 3.07 \\
\midrule
\multicolumn{2}{c|}{Joint Socio-Temporal} &  $\text{ADE}_5\downarrow$ & $\text{FDE}_5\downarrow$ & $\text{ADE}_{10}\downarrow$ & $\text{FDE}_{10}\downarrow$\\ \midrule%
\multicolumn{2}{l|}{Ours w/o semantic map}  & 1.97 & 4.21 & 1.58 & 3.14 \\
\multicolumn{2}{l|}{Ours w/o joint latent}  & 1.95 & 3.98 & 1.50 & 2.92 \\
\multicolumn{2}{l|}{Ours w/o AA attention} & 2.02 & 4.29 & 1.55 & 2.91 \\
\multicolumn{2}{l|}{Ours w/ agent encoding} & 2.01 & 4.28 & 1.63 & 3.11 \\
\multicolumn{2}{l|}{Ours (\mname)} & \textbf{1.86} & \textbf{3.89} & \textbf{1.45} & \textbf{2.86}\\
\bottomrule
\end{tabular}
\vspace{1.5mm}
\caption{\textbf{Ablation studies} on the nuScenes dataset. ``TF'' means Transformer and ``AA Attention'' denotes agent-aware attention.}
\label{table:nuscenes_abl}
\vspace{-3mm}
\end{table}

\vspace{2mm}
\noindent\textbf{Ablation Studies.}
We further perform extensive ablation studies on ETH/UCY and nuScenes to investigate the contribution of key technical components in our method. The first ablation study explores variants of our method that use separate social and temporal models to replace our joint socio-temporal model, \mname, in our multi-agent prediction framework. We choose GCN~\cite{kipf2016semi} or Transformer (TF) as the social model, and LSTM or Transformer as the temporal model. In total, there are 4 ($2\times2$) combinations of social and temporal models. The ablation results are summarized in the first group of Table~\ref{table:eth_abl} and~\ref{table:nuscenes_abl}. It is evident that all combinations of separate social and temporal models lead to inferior performance compared to our method which models the social and temporal dimensions jointly. 

The second ablation study investigates the role of (1) joint latent intent modeling, (2) agent-aware attention, and (3) semantic maps, and we denote the corresponding variants as ``w/o joint latent'', ``w/o AA attention'', and ``w/o semantic map''. We further test a variant ``w/ agent encoding'' where we replace agent-aware attention with agent encoding. The results are reported in the second group of Table~\ref{table:eth_abl} and~\ref{table:nuscenes_abl}. We can see that all variants lead to considerably worse performance compared to our full method. In particular, the variants ``w/o AA attention'' and ``w/ agent encoding'' result in pronounced performance drop, which indicates that agent-aware attention is essential in our method and alternatives like agent encoding are not effective.

\begin{figure}
    \centering
    \includegraphics[width=\linewidth]{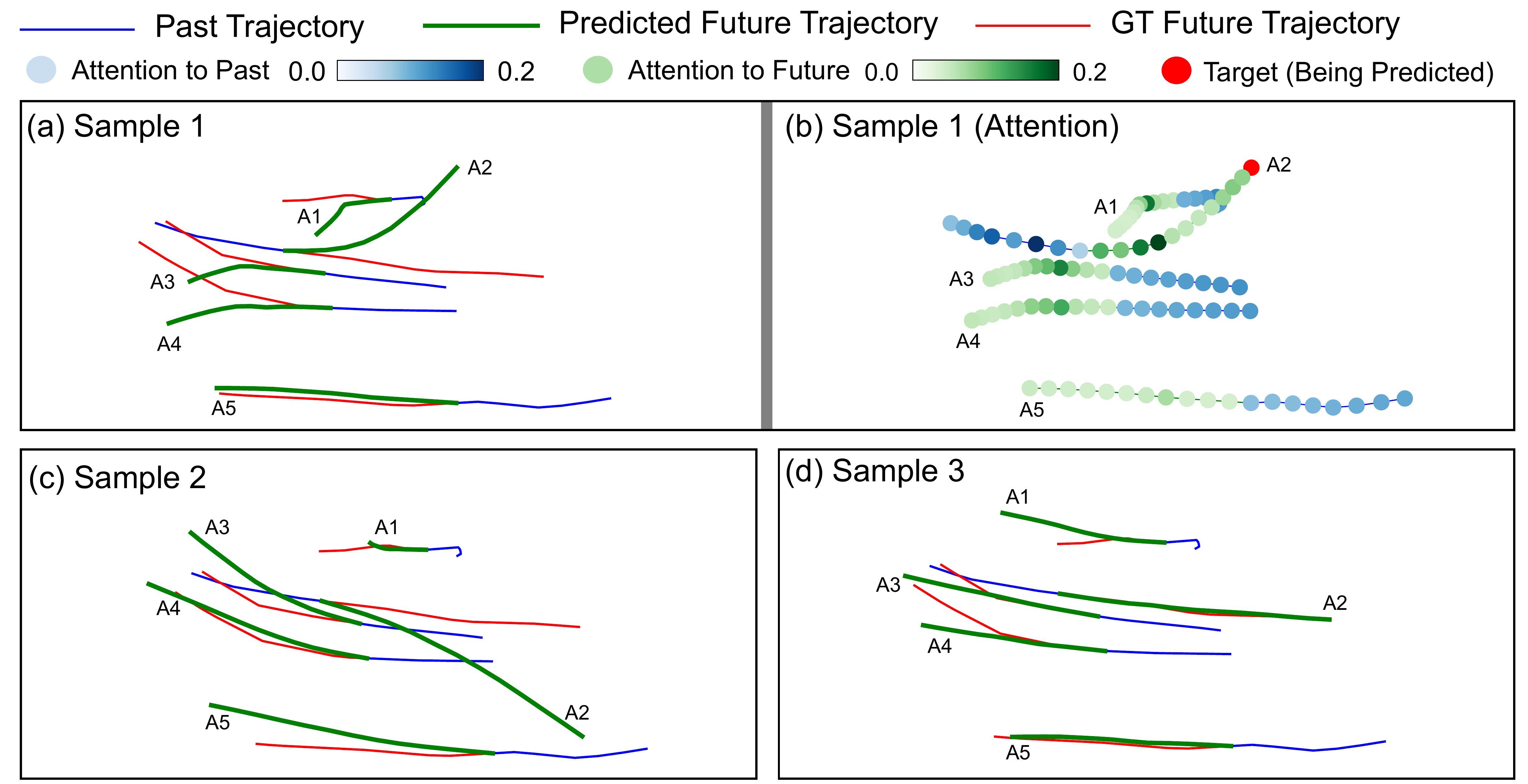}
    \caption{\textbf{(a,c,d)} Three samples of forecasted multi-agent futures (green) via our method, which exhibit social behaviors like following (A3 \& A4) and collision avoidance (A1 \& A2 in (a), A2 \& A3 in (c)). \textbf{(b)} Attention visualization for sample 1. When predicting the target (red), the model pays more attention (darker color) to key timesteps (turning point) of adjacent agents and spreads out attention to the target's past timesteps to reason about dynamics.}
    \label{fig:vis}
    \vspace{-3mm}
\end{figure}

\vspace{2mm}
\noindent\textbf{Trajectory Visualization.}
Fig.~\ref{fig:vis}\,(a,c,d) shows three samples of forecasted multi-agent futures of the same scene via our method. We can see that the samples correspond to different modes of socially-aware and non-colliding trajectories, and exhibit behaviors like following (A3 \& A4) and collision avoidance (A1 \& A2 in (a), A2 \& A3 in (c)). Fig.~\ref{fig:vis}\,(b) visualizes the attention of sample 1 and shows that, when predicting the target (red), the model pays more attention to key timesteps (turning point) of adjacent agents and also spreads out attention to the target's past timesteps to reason about the dynamics and curvature of its trajectory. More attention visualization can be found in Appendix~\ref{sec:supp_vis_attn}.

\section{Conclusion}
In this paper, we proposed a new Transformer, \mname, that can simultaneously model the time and social dimensions of multi-agent trajectories using a sequence representation. To preserve agent identities in the sequence, we proposed a novel agent-aware attention mechanism that can attend to features of the same agent differently than features of other agents. Based on \mname, we presented a stochastic multi-agent trajectory prediction framework that jointly models the latent intent of all agents to produce diverse and socially-aware multi-agent future trajectories. Experiments demonstrated that our method substantially improved state-of-the-art performance on challenging pedestrian and autonomous driving datasets.

\vspace{2mm}
\noindent\textbf{Acknowledgments.} This work is supported by the Qualcomm Innovation Fellowship.

{\small
\bibliographystyle{ieee_fullname}
\bibliography{ref}

\begin{thebibliography}{10}\itemsep=-1pt

\bibitem{alahi2016social}
Alexandre Alahi, Kratarth Goel, Vignesh Ramanathan, Alexandre Robicquet, Li
  Fei-Fei, and Silvio Savarese.
\newblock Social lstm: Human trajectory prediction in crowded spaces.
\newblock In {\em Proceedings of the IEEE conference on computer vision and
  pattern recognition}, pages 961--971, 2016.

\bibitem{bahdanau2014neural}
Dzmitry Bahdanau, Kyunghyun Cho, and Yoshua Bengio.
\newblock Neural machine translation by jointly learning to align and
  translate.
\newblock {\em arXiv preprint arXiv:1409.0473}, 2014.

\bibitem{caesar2020nuscenes}
Holger Caesar, Varun Bankiti, Alex~H Lang, Sourabh Vora, Venice~Erin Liong,
  Qiang Xu, Anush Krishnan, Yu Pan, Giancarlo Baldan, and Oscar Beijbom.
\newblock nuscenes: A multimodal dataset for autonomous driving.
\newblock In {\em Proceedings of the IEEE/CVF conference on computer vision and
  pattern recognition}, pages 11621--11631, 2020.

\bibitem{carion2020end}
Nicolas Carion, Francisco Massa, Gabriel Synnaeve, Nicolas Usunier, Alexander
  Kirillov, and Sergey Zagoruyko.
\newblock End-to-end object detection with transformers.
\newblock In {\em European Conference on Computer Vision}, pages 213--229.
  Springer, 2020.

\bibitem{chai2020multipath}
Yuning Chai, Benjamin Sapp, Mayank Bansal, and Dragomir Anguelov.
\newblock Multipath: Multiple probabilistic anchor trajectory hypotheses for
  behavior prediction.
\newblock In {\em Conference on Robot Learning}, pages 86--99. PMLR, 2020.

\bibitem{cho2014learning}
Kyunghyun Cho, Bart Van~Merri{\"e}nboer, Caglar Gulcehre, Dzmitry Bahdanau,
  Fethi Bougares, Holger Schwenk, and Yoshua Bengio.
\newblock Learning phrase representations using rnn encoder-decoder for
  statistical machine translation.
\newblock {\em arXiv preprint arXiv:1406.1078}, 2014.

\bibitem{chung2014empirical}
Junyoung Chung, Caglar Gulcehre, KyungHyun Cho, and Yoshua Bengio.
\newblock Empirical evaluation of gated recurrent neural networks on sequence
  modeling.
\newblock {\em arXiv preprint arXiv:1412.3555}, 2014.

\bibitem{cui2019multimodal}
Henggang Cui, Vladan Radosavljevic, Fang-Chieh Chou, Tsung-Han Lin, Thi Nguyen,
  Tzu-Kuo Huang, Jeff Schneider, and Nemanja Djuric.
\newblock Multimodal trajectory predictions for autonomous driving using deep
  convolutional networks.
\newblock In {\em 2019 International Conference on Robotics and Automation
  (ICRA)}, pages 2090--2096. IEEE, 2019.

\bibitem{devlin2018bert}
Jacob Devlin, Ming-Wei Chang, Kenton Lee, and Kristina Toutanova.
\newblock Bert: Pre-training of deep bidirectional transformers for language
  understanding.
\newblock {\em arXiv preprint arXiv:1810.04805}, 2018.

\bibitem{dosovitskiy2020image}
Alexey Dosovitskiy, Lucas Beyer, Alexander Kolesnikov, Dirk Weissenborn,
  Xiaohua Zhai, Thomas Unterthiner, Mostafa Dehghani, Matthias Minderer, Georg
  Heigold, Sylvain Gelly, et~al.
\newblock An image is worth 16x16 words: Transformers for image recognition at
  scale.
\newblock {\em arXiv preprint arXiv:2010.11929}, 2020.

\bibitem{fragkiadaki2015recurrent}
Katerina Fragkiadaki, Sergey Levine, Panna Felsen, and Jitendra Malik.
\newblock Recurrent network models for human dynamics.
\newblock In {\em Proceedings of the IEEE International Conference on Computer
  Vision}, pages 4346--4354, 2015.

\bibitem{giuliari2020transformer}
Francesco Giuliari, Irtiza Hasan, Marco Cristani, and Fabio Galasso.
\newblock Transformer networks for trajectory forecasting.
\newblock {\em ICPR}, 2020.

\bibitem{goodfellow2014generative}
Ian~J Goodfellow, Jean Pouget-Abadie, Mehdi Mirza, Bing Xu, David Warde-Farley,
  Sherjil Ozair, Aaron Courville, and Yoshua Bengio.
\newblock Generative adversarial networks.
\newblock {\em arXiv preprint arXiv:1406.2661}, 2014.

\bibitem{guan2020generative}
Jiaqi Guan, Ye Yuan, Kris~M Kitani, and Nicholas Rhinehart.
\newblock Generative hybrid representations for activity forecasting with
  no-regret learning.
\newblock In {\em Proceedings of the IEEE/CVF Conference on Computer Vision and
  Pattern Recognition}, pages 173--182, 2020.

\bibitem{gupta2018social}
Agrim Gupta, Justin Johnson, Li Fei-Fei, Silvio Savarese, and Alexandre Alahi.
\newblock Social gan: Socially acceptable trajectories with generative
  adversarial networks.
\newblock In {\em Proceedings of the IEEE Conference on Computer Vision and
  Pattern Recognition}, pages 2255--2264, 2018.

\bibitem{helbing1995social}
Dirk Helbing and Peter Molnar.
\newblock Social force model for pedestrian dynamics.
\newblock {\em Physical review E}, 51(5):4282, 1995.

\bibitem{hochreiter1997long}
Sepp Hochreiter and J{\"u}rgen Schmidhuber.
\newblock Long short-term memory.
\newblock {\em Neural computation}, 9(8):1735--1780, 1997.

\bibitem{huang2019stgat}
Yingfan Huang, Huikun Bi, Zhaoxin Li, Tianlu Mao, and Zhaoqi Wang.
\newblock Stgat: Modeling spatial-temporal interactions for human trajectory
  prediction.
\newblock In {\em Proceedings of the IEEE/CVF International Conference on
  Computer Vision}, pages 6272--6281, 2019.

\bibitem{ivanovic2019trajectron}
Boris Ivanovic and Marco Pavone.
\newblock The trajectron: Probabilistic multi-agent trajectory modeling with
  dynamic spatiotemporal graphs.
\newblock In {\em Proceedings of the IEEE/CVF International Conference on
  Computer Vision}, pages 2375--2384, 2019.

\bibitem{kingma2014adam}
Diederik~P Kingma and Jimmy Ba.
\newblock Adam: A method for stochastic optimization.
\newblock {\em arXiv preprint arXiv:1412.6980}, 2014.

\bibitem{kingma2013auto}
Diederik~P Kingma and Max Welling.
\newblock Auto-encoding variational bayes.
\newblock {\em arXiv preprint arXiv:1312.6114}, 2013.

\bibitem{kipf2018neural}
Thomas Kipf, Ethan Fetaya, Kuan-Chieh Wang, Max Welling, and Richard Zemel.
\newblock Neural relational inference for interacting systems.
\newblock In {\em International Conference on Machine Learning}, pages
  2688--2697. PMLR, 2018.

\bibitem{kipf2016semi}
Thomas~N Kipf and Max Welling.
\newblock Semi-supervised classification with graph convolutional networks.
\newblock {\em arXiv preprint arXiv:1609.02907}, 2016.

\bibitem{kocabas2020vibe}
Muhammed Kocabas, Nikos Athanasiou, and Michael~J Black.
\newblock Vibe: Video inference for human body pose and shape estimation.
\newblock In {\em Proceedings of the IEEE/CVF Conference on Computer Vision and
  Pattern Recognition}, pages 5253--5263, 2020.

\bibitem{kosaraju2019social}
Vineet Kosaraju, Amir Sadeghian, Roberto Mart{\'\i}n-Mart{\'\i}n, Ian~D Reid,
  Hamid Rezatofighi, and Silvio Savarese.
\newblock Social-bigat: multimodal trajectory forecasting using bicycle-gan and
  graph attention networks.
\newblock In {\em Advances in Neural Information Processing Systems 2019}.
  Neural Information Processing Systems (NIPS), 2019.

\bibitem{lan2019albert}
Zhenzhong Lan, Mingda Chen, Sebastian Goodman, Kevin Gimpel, Piyush Sharma, and
  Radu Soricut.
\newblock Albert: A lite bert for self-supervised learning of language
  representations.
\newblock {\em arXiv preprint arXiv:1909.11942}, 2019.

\bibitem{lee2017desire}
Namhoon Lee, Wongun Choi, Paul Vernaza, Christopher~B Choy, Philip~HS Torr, and
  Manmohan Chandraker.
\newblock Desire: Distant future prediction in dynamic scenes with interacting
  agents.
\newblock In {\em Proceedings of the IEEE Conference on Computer Vision and
  Pattern Recognition}, pages 336--345, 2017.

\bibitem{lerner2007crowds}
Alon Lerner, Yiorgos Chrysanthou, and Dani Lischinski.
\newblock Crowds by example.
\newblock In {\em Computer graphics forum}, volume~26, pages 655--664. Wiley
  Online Library, 2007.

\bibitem{li2020evolvegraph}
Jiachen Li, Fan Yang, Masayoshi Tomizuka, and Chiho Choi.
\newblock Evolvegraph: Multi-agent trajectory prediction with dynamic
  relational reasoning.
\newblock {\em Advances in Neural Information Processing Systems}, 33, 2020.

\bibitem{li2020end}
Lingyun~Luke Li, Bin Yang, Ming Liang, Wenyuan Zeng, Mengye Ren, Sean Segal,
  and Raquel Urtasun.
\newblock End-to-end contextual perception and prediction with interaction
  transformer.
\newblock {\em arXiv preprint arXiv:2008.05927}, 2020.

\bibitem{li2015gated}
Yujia Li, Daniel Tarlow, Marc Brockschmidt, and Richard Zemel.
\newblock Gated graph sequence neural networks.
\newblock {\em arXiv preprint arXiv:1511.05493}, 2015.

\bibitem{luong2015effective}
Minh-Thang Luong, Hieu Pham, and Christopher~D Manning.
\newblock Effective approaches to attention-based neural machine translation.
\newblock {\em arXiv preprint arXiv:1508.04025}, 2015.

\bibitem{ma2020diverse}
Yecheng~Jason Ma, Jeevana~Priya Inala, Dinesh Jayaraman, and Osbert Bastani.
\newblock Diverse sampling for normalizing flow based trajectory forecasting.
\newblock {\em arXiv preprint arXiv:2011.15084}, 2020.

\bibitem{mangalam2020not}
Karttikeya Mangalam, Harshayu Girase, Shreyas Agarwal, Kuan-Hui Lee, Ehsan
  Adeli, Jitendra Malik, and Adrien Gaidon.
\newblock It is not the journey but the destination: Endpoint conditioned
  trajectory prediction.
\newblock In {\em European Conference on Computer Vision}, pages 759--776.
  Springer, 2020.

\bibitem{miao2015eesen}
Yajie Miao, Mohammad Gowayyed, and Florian Metze.
\newblock Eesen: End-to-end speech recognition using deep rnn models and
  wfst-based decoding.
\newblock In {\em 2015 IEEE Workshop on Automatic Speech Recognition and
  Understanding (ASRU)}, pages 167--174. IEEE, 2015.

\bibitem{morton2016analysis}
Jeremy Morton, Tim~A Wheeler, and Mykel~J Kochenderfer.
\newblock Analysis of recurrent neural networks for probabilistic modeling of
  driver behavior.
\newblock {\em IEEE Transactions on Intelligent Transportation Systems},
  18(5):1289--1298, 2016.

\bibitem{paszke2019pytorch}
Adam Paszke, Sam Gross, Francisco Massa, Adam Lerer, James Bradbury, Gregory
  Chanan, Trevor Killeen, Zeming Lin, Natalia Gimelshein, Luca Antiga, et~al.
\newblock Pytorch: An imperative style, high-performance deep learning library.
\newblock {\em arXiv preprint arXiv:1912.01703}, 2019.

\bibitem{pellegrini2009you}
Stefano Pellegrini, Andreas Ess, Konrad Schindler, and Luc Van~Gool.
\newblock You'll never walk alone: Modeling social behavior for multi-target
  tracking.
\newblock In {\em 2009 IEEE 12th International Conference on Computer Vision},
  pages 261--268. IEEE, 2009.

\bibitem{phan2020covernet}
Tung Phan-Minh, Elena~Corina Grigore, Freddy~A Boulton, Oscar Beijbom, and
  Eric~M Wolff.
\newblock Covernet: Multimodal behavior prediction using trajectory sets.
\newblock In {\em Proceedings of the IEEE/CVF Conference on Computer Vision and
  Pattern Recognition}, pages 14074--14083, 2020.

\bibitem{rezende2015variational}
Danilo Rezende and Shakir Mohamed.
\newblock Variational inference with normalizing flows.
\newblock In {\em International Conference on Machine Learning}, pages
  1530--1538. PMLR, 2015.

\bibitem{rhinehart2018r2p2}
Nicholas Rhinehart, Kris~M Kitani, and Paul Vernaza.
\newblock R2p2: A reparameterized pushforward policy for diverse, precise
  generative path forecasting.
\newblock In {\em Proceedings of the European Conference on Computer Vision
  (ECCV)}, pages 772--788, 2018.

\bibitem{rhinehart2019precog}
Nicholas Rhinehart, Rowan McAllister, Kris Kitani, and Sergey Levine.
\newblock Precog: Prediction conditioned on goals in visual multi-agent
  settings.
\newblock In {\em Proceedings of the IEEE/CVF International Conference on
  Computer Vision}, pages 2821--2830, 2019.

\bibitem{rudenko2020human}
Andrey Rudenko, Luigi Palmieri, Michael Herman, Kris~M Kitani, Dariu~M Gavrila,
  and Kai~O Arras.
\newblock Human motion trajectory prediction: A survey.
\newblock {\em The International Journal of Robotics Research}, 39(8):895--935,
  2020.

\bibitem{sadeghian2019sophie}
Amir Sadeghian, Vineet Kosaraju, Ali Sadeghian, Noriaki Hirose, Hamid
  Rezatofighi, and Silvio Savarese.
\newblock Sophie: An attentive gan for predicting paths compliant to social and
  physical constraints.
\newblock In {\em Proceedings of the IEEE/CVF Conference on Computer Vision and
  Pattern Recognition}, pages 1349--1358, 2019.

\bibitem{salzmann2020trajectron++}
Tim Salzmann, Boris Ivanovic, Punarjay Chakravarty, and Marco Pavone.
\newblock Trajectron++: Dynamically-feasible trajectory forecasting with
  heterogeneous data.
\newblock {\em arXiv preprint arXiv:2001.03093}, 2020.

\bibitem{tang2019multiple}
Yichuan~Charlie Tang and Ruslan Salakhutdinov.
\newblock Multiple futures prediction.
\newblock {\em arXiv preprint arXiv:1911.00997}, 2019.

\bibitem{vaswani2017attention}
Ashish Vaswani, Noam Shazeer, Niki Parmar, Jakob Uszkoreit, Llion Jones,
  Aidan~N Gomez, {\L}ukasz Kaiser, and Illia Polosukhin.
\newblock Attention is all you need.
\newblock In {\em Proceedings of the 31st International Conference on Neural
  Information Processing Systems}, pages 6000--6010, 2017.

\bibitem{vemula2018social}
Anirudh Vemula, Katharina Muelling, and Jean Oh.
\newblock Social attention: Modeling attention in human crowds.
\newblock In {\em 2018 IEEE international Conference on Robotics and Automation
  (ICRA)}, pages 4601--4607. IEEE, 2018.

\bibitem{wang2007gaussian}
Jack~M Wang, David~J Fleet, and Aaron Hertzmann.
\newblock Gaussian process dynamical models for human motion.
\newblock {\em IEEE transactions on pattern analysis and machine intelligence},
  30(2):283--298, 2007.

\bibitem{wang2020end}
Yuqing Wang, Zhaoliang Xu, Xinlong Wang, Chunhua Shen, Baoshan Cheng, Hao Shen,
  and Huaxia Xia.
\newblock End-to-end video instance segmentation with transformers.
\newblock {\em arXiv preprint arXiv:2011.14503}, 2020.

\bibitem{weng2020joint}
Xinshuo Weng, Ye Yuan, and Kris Kitani.
\newblock Joint 3d tracking and forecasting with graph neural network and
  diversity sampling.
\newblock {\em arXiv preprint arXiv:2003.07847}, 2020.

\bibitem{xiong2018microsoft}
Wayne Xiong, Lingfeng Wu, Fil Alleva, Jasha Droppo, Xuedong Huang, and Andreas
  Stolcke.
\newblock The microsoft 2017 conversational speech recognition system.
\newblock In {\em 2018 IEEE international conference on acoustics, speech and
  signal processing (ICASSP)}, pages 5934--5938. IEEE, 2018.

\bibitem{xu2015show}
Kelvin Xu, Jimmy Ba, Ryan Kiros, Kyunghyun Cho, Aaron Courville, Ruslan
  Salakhudinov, Rich Zemel, and Yoshua Bengio.
\newblock Show, attend and tell: Neural image caption generation with visual
  attention.
\newblock In {\em International conference on machine learning}, pages
  2048--2057. PMLR, 2015.

\bibitem{yang2019xlnet}
Zhilin Yang, Zihang Dai, Yiming Yang, Jaime Carbonell, Ruslan Salakhutdinov,
  and Quoc~V Le.
\newblock Xlnet: Generalized autoregressive pretraining for language
  understanding.
\newblock {\em arXiv preprint arXiv:1906.08237}, 2019.

\bibitem{yu2020spatio}
Cunjun Yu, Xiao Ma, Jiawei Ren, Haiyu Zhao, and Shuai Yi.
\newblock Spatio-temporal graph transformer networks for pedestrian trajectory
  prediction.
\newblock In {\em European Conference on Computer Vision}, pages 507--523.
  Springer, 2020.

\bibitem{yuan20183d}
Ye Yuan and Kris Kitani.
\newblock 3d ego-pose estimation via imitation learning.
\newblock In {\em Proceedings of the European Conference on Computer Vision
  (ECCV)}, pages 735--750, 2018.

\bibitem{yuan2019diverse}
Ye Yuan and Kris Kitani.
\newblock Diverse trajectory forecasting with determinantal point processes.
\newblock {\em arXiv preprint arXiv:1907.04967}, 2019.

\bibitem{yuan2019ego}
Ye Yuan and Kris Kitani.
\newblock Ego-pose estimation and forecasting as real-time pd control.
\newblock In {\em Proceedings of the IEEE/CVF International Conference on
  Computer Vision}, pages 10082--10092, 2019.

\bibitem{yuan2020dlow}
Ye Yuan and Kris Kitani.
\newblock Dlow: Diversifying latent flows for diverse human motion prediction.
\newblock In {\em European Conference on Computer Vision}, pages 346--364.
  Springer, 2020.

\bibitem{yuan2020residual}
Ye Yuan and Kris Kitani.
\newblock Residual force control for agile human behavior imitation and
  extended motion synthesis.
\newblock In {\em Advances in Neural Information Processing Systems}, 2020.

\bibitem{zhang2019sr}
Pu Zhang, Wanli Ouyang, Pengfei Zhang, Jianru Xue, and Nanning Zheng.
\newblock Sr-lstm: State refinement for lstm towards pedestrian trajectory
  prediction.
\newblock In {\em Proceedings of the IEEE/CVF Conference on Computer Vision and
  Pattern Recognition}, pages 12085--12094, 2019.

\bibitem{zhao2019multi}
Tianyang Zhao, Yifei Xu, Mathew Monfort, Wongun Choi, Chris Baker, Yibiao Zhao,
  Yizhou Wang, and Ying~Nian Wu.
\newblock Multi-agent tensor fusion for contextual trajectory prediction.
\newblock In {\em Proceedings of the IEEE/CVF Conference on Computer Vision and
  Pattern Recognition}, pages 12126--12134, 2019.

\end{thebibliography}
}

\onecolumn
\appendix

\section{Handling a Time-Varying Number of Agents}
\begin{figure*}[h]
    \centering
    \includegraphics[width=0.95\textwidth]{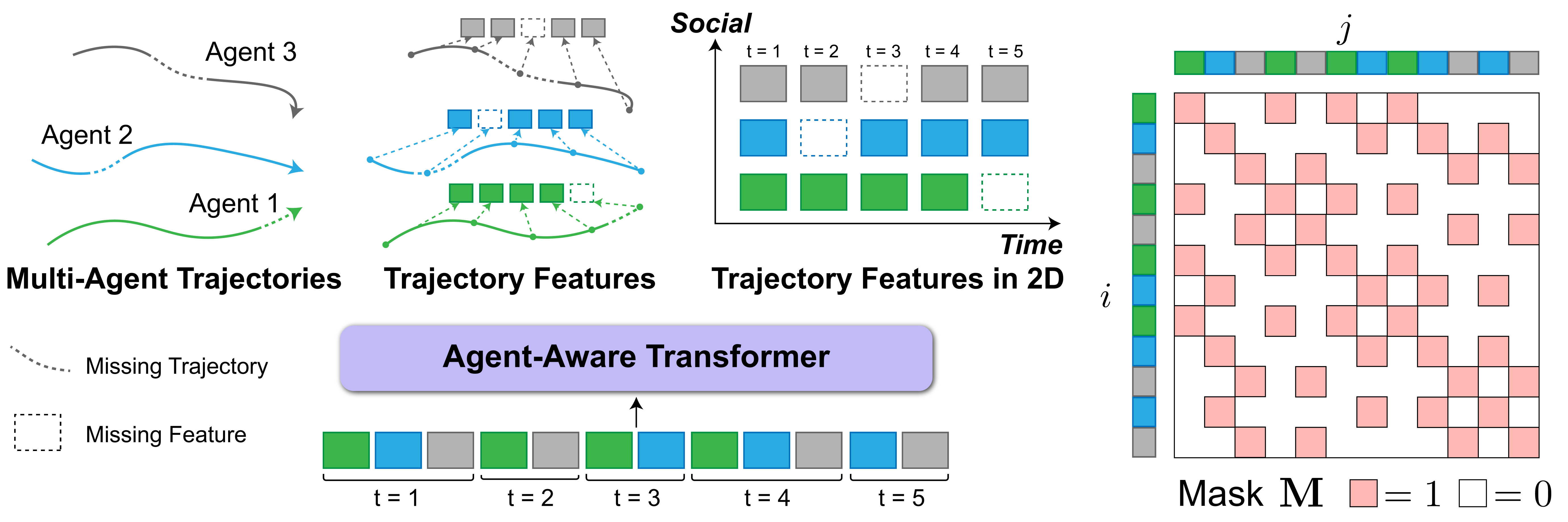}
    \caption{Our method can naturally handle a time-varying number of agents because of the flexible sequence representation of multi-agent trajectories. We can simply remove the trajectory features of missing agents at each timestep from the sequence. The mask $\mb{M}$ of the example sequence (when applying self-attention) is computed based on the agreement of agent identity between each query and key.}
    \label{fig:var_agent}
    \vspace{-2mm}
\end{figure*}

For clarity and ease of exposition, we assume the number of agents remains the same across timesteps in the main paper. However, this assumption is not necessary, and our method can easily generalize to use cases where the number of agents changes over time due to agents going out of the scene or being missed by detection. We illustrate how to apply our method to such cases in Fig.~\ref{fig:var_agent}. Owning to the flexible sequence representation we employ for multi-agent trajectories, we can simply remove the features of missing agents at each timestep from the sequence. The reason why we do not need to fill the missing features is that our method uses time encoding to preserve time information, unlike RNNs which have to use recurrence to encode timesteps and thus necessitate the features of all timesteps. As the number of agents is no longer $N$ for all timesteps, the computation of the mask $\mb{M}$ in agent-aware attention needs to be changed accordingly:
\begin{equation}
M_{ij} = \mathbbm{1}  (\text{Agent}(i) = \text{Agent}(j))
\end{equation}
where Agent($\cdot$) extracts the agent index of a query/key and $\mathbbm{1}(\cdot)$ denotes the indicator function. An example of mask $\mb{M}$ is shown in Fig.~\ref{fig:var_agent}\,(Right).

\section{Additional Implementation Details}
\label{sec:supp_details}
\noindent\textbf{Encoding Semantic Maps.}
The semantic map $\mb{I}_n \in \mathbb{R}^{H\times W \times C}$ for each agent $n$ has spatial dimensions $(100, 100)$ with 3 meters between adjacent pixels. It has $C = 3$ channels annotating drivable areas, road dividers, and lane dividers obtained using the official nuScenes software development kit. Since the semantic map is relatively easy to parse, we use a simple hand-designed CNN to extract visual features $\mb{v}_n$ from it. In particular, the CNN has four convolutional layers with channels (32, 32, 32, 1), kernel size (5, 5, 5, 3), and strides (2, 2, 1, 1). A final linear layer is used to obtain a 32-dimensional feature.

\vspace{2mm}
\noindent\textbf{Training Trajectory Sampler.}
The scaling factor $\sigma_d$ in the trajectory sampler loss $L_{samp}$ (Eq.~\eqref{eq:samp_loss} in the main paper) is set to 5 for ETH/UCY and 20 for nuScenes. We clip the maximum value of the KL term in $L_{samp}$ down to 2. We train the trajectory sampler using the Adam optimizer~\cite{kingma2014adam} for 50 epochs on ETH/UCY and nuScenes. We use an initial learning rate of $10^{-4}$ and halve the learning rate every 5 epochs.

\vspace{2mm}
\noindent\textbf{Ablation Study Details.}
We first provide details for the ablation study of separate social and temporal models (first group of Table~\ref{table:eth_abl} and \ref{table:nuscenes_abl} in the main paper). We first use a temporal model (LSTM or Transformer) to extract the temporal feature of each agent and then apply a social model (GCN~\cite{kipf2016semi} or Transformer) over the temporal features to obtain social features for each agent; final trajectories are decoded from the social features using either an LSTM or Transformer. For the GCN, we use two graph convolutional layers with channels (256, 256) and residual connections within each layer. The hidden dimensions of the LSTMs are set to 256. The Transformers have two layers with key/query dimensions 256 and 8 heads; the feedforward layer has 512 hidden units, and the dropout ratio is 0.1. We use the positional encoding~\cite{vaswani2017attention} for the temporal Transformer but not for the social Transformer as agents are permutation-invariant.

Next, we provide details for the ablation study of each key technical component (second group of Table~\ref{table:eth_abl} and \ref{table:nuscenes_abl} in the main paper). For the variant without joint latent modeling (``w/o joint latent''), we append the latent codes to the trajectory sequence after the \mname\ decoder instead of before the decoder. In this way, the latent code of one agent will not affect the future trajectory of another agent. For the variant without the agent-aware attention (``w/o AA attention''), we replace our agent-aware attention with standard scaled dot-product attention used in the original transformer~\cite{vaswani2017attention}. For the variant with agent encoding (``w/ agent encoding''), in addition to removing the agent aware attention, we also append an agent encoding to each element in the trajectory sequence. The agent encoding is computed similarly as the positional encoding~\cite{vaswani2017attention} but uses the agent index instead of the position index. For the variant without semantic maps (``w/o semantic map''), we simply do not append any visual features extracted from the semantic maps to the trajectory sequence.

\vspace{2mm}
\noindent\textbf{Other Details.}
Our models are implemented using PyTorch~\cite{paszke2019pytorch} and are trained with a single NVIDIA RTX 2080 Ti and standard CPUs. The training time is approximately one day for each dataset in ETH/UCY and three days for nuScenes.

\section{Additional Attention Visualization}
\label{sec:supp_vis_attn}
\begin{figure*}[h]
\vspace{-3mm}
    \centering
    \includegraphics[width=\linewidth]{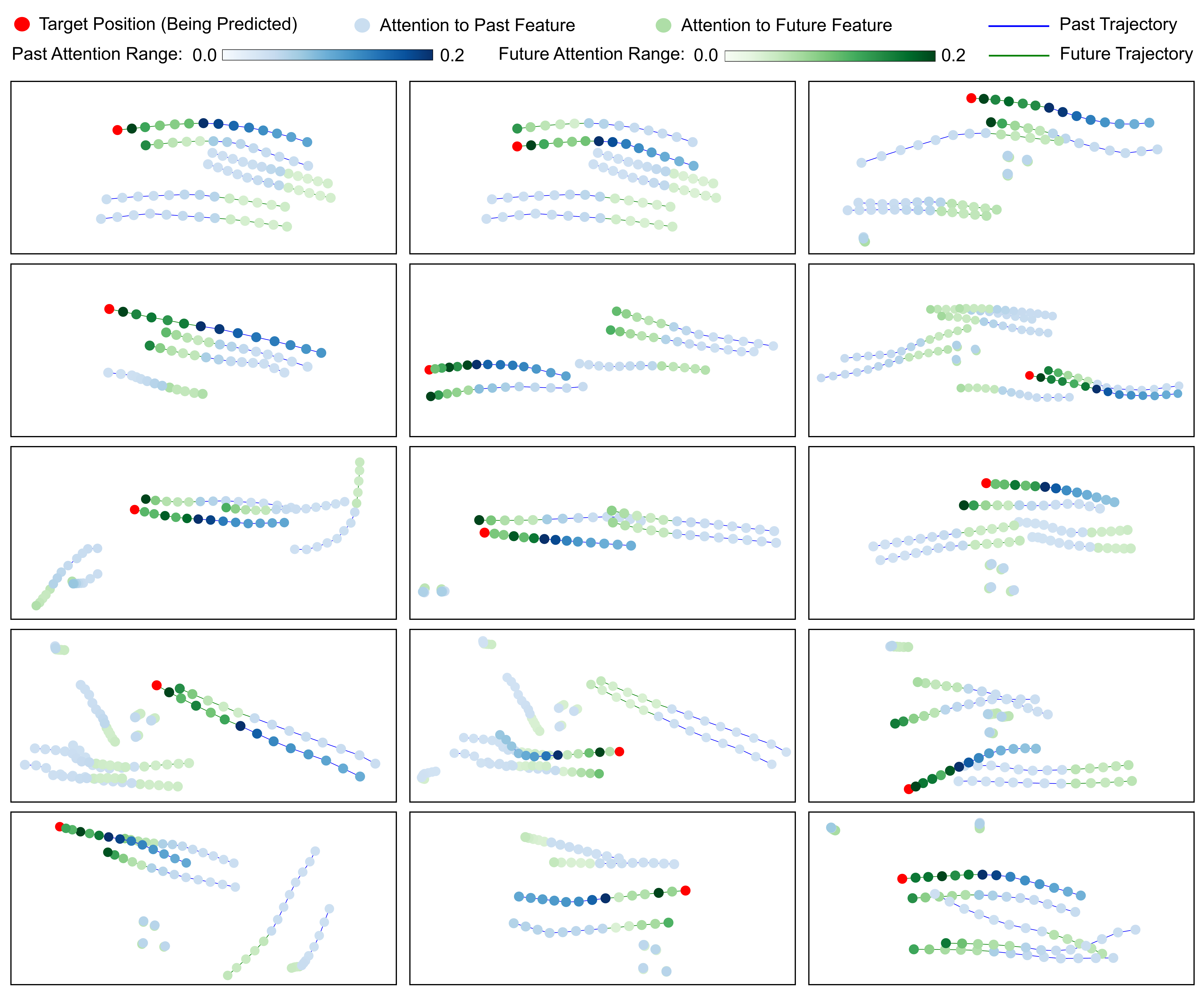}
    \caption{\textbf{Attention Visualization on ETH/UCY.} We plot the attention to past (blue) and future (green) trajectory features of all agents when inferring a target position (red). Darker color means higher attention. When predicting the target future position of an agent, the model pays more attention to the agent's own trajectories and recent timesteps, and it also attends more to nearby agents than distant agents.}
    \label{fig:vis_attn_supp}
    \vspace{-3mm}
\end{figure*}

As discussed in the main paper, our method can attend to any agent at any previous timestep when predicting the future position of an agent. Here, we provide more visualization of the attention in Fig.~\ref{fig:vis_attn_supp} to understand the behavior of our model. Across all the examples, it is evident that when predicting the target future position of an agent, the model pays more attention to the agent's own trajectories and recent timesteps, and it also attends more to nearby agents than distant agents.

\section{Trajectory Sample Visualization}
\begin{figure*}[h]
    \centering
    \includegraphics[width=\linewidth]{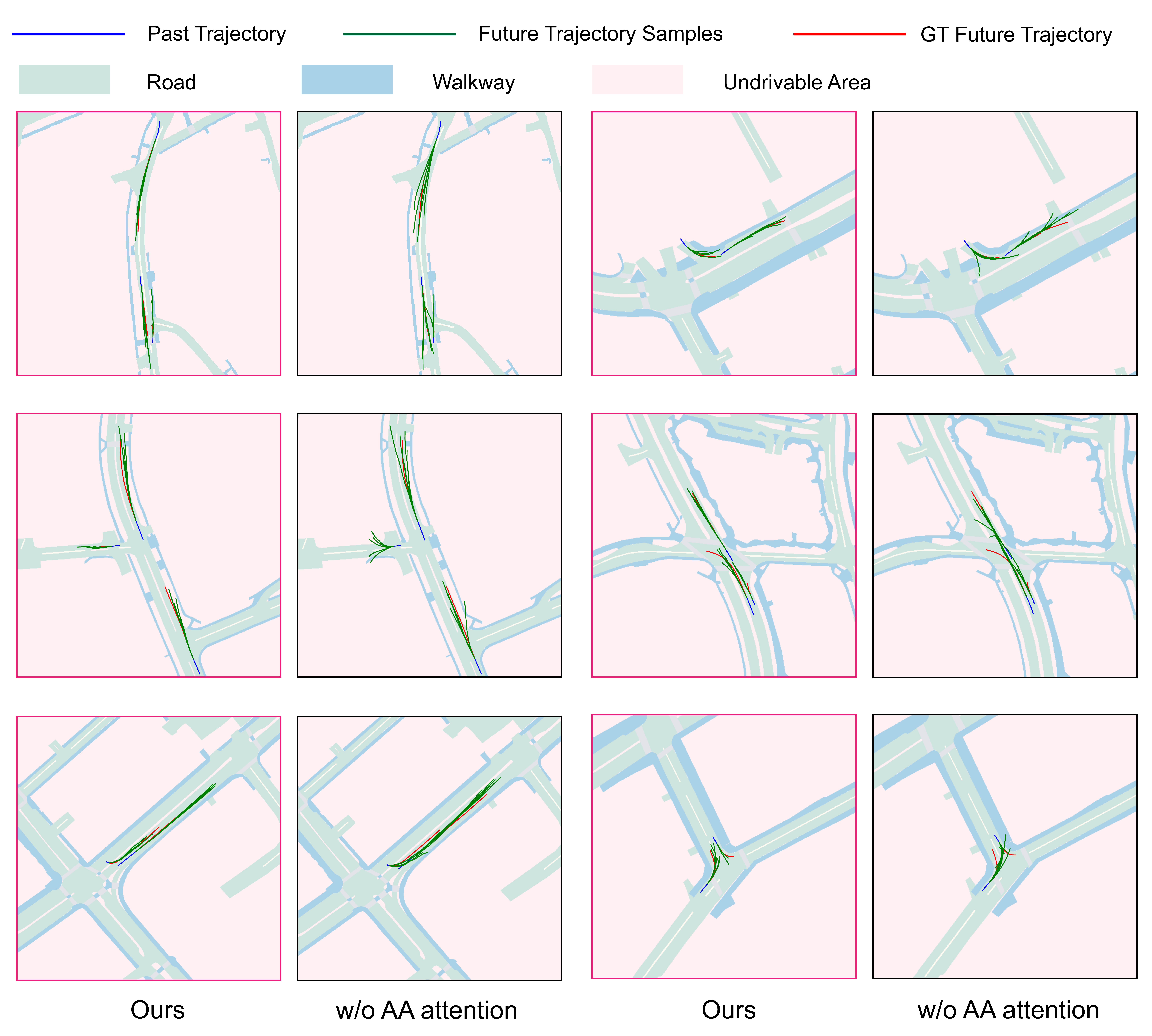}
    \caption{\textbf{Trajectory Sample Visualization on nuScenes.} We compare our method against the variant without agent-aware attention (w/o AA attention). The future trajectory samples produced by our method using agent-aware attention cover the ground truth (GT) future trajectories significantly better. Our method also produces much fewer implausible trajectories such as those going out of the road.}
    \label{fig:vis_nus}
    \vspace{-2mm}
\end{figure*}

 To demonstrate the importance of agent-aware attention, we also provide qualitative comparisons of our method against the variant without agent-aware attention (w/o AA attention) on the nuScenes dataset in Fig.~\ref{fig:vis_nus}. We can observe that the future trajectory samples produced by our method using agent-aware attention cover the ground truth (GT) future trajectories significantly better. Our method also produces much fewer implausible trajectories such as those going out of the road.

\end{document}